\documentclass[final,3p,times,authoryear,review,preprint,onecolumn]{elsarticle}

\usepackage{amssymb}
\usepackage{apalike}
\usepackage[longtable]{multirow}
\usepackage{longtable}
\usepackage{array}
\usepackage[labelfont=bf]{caption}
\usepackage{colortbl}
\usepackage{amsmath}
\usepackage{graphicx}
\usepackage{ragged2e}
\usepackage[hidelinks]{hyperref}
\hypersetup{
  colorlinks   = true, 
  urlcolor     = blue, 
  linkcolor    = blue, 
  citecolor   = blue 
}
\usepackage{multirow}
\usepackage{float}
\usepackage{soul}
\usepackage{caption}
\usepackage{textcomp}
\usepackage{multicol}
\usepackage{multirow}
\usepackage{interval}
\usepackage{adjustbox}
\usepackage{graphicx}%
\usepackage{multirow}%
\usepackage{amsthm}%
\usepackage{mathrsfs}%
\usepackage{subcaption} 
\usepackage{xcolor}%

\usepackage{textcomp}%
\usepackage{manyfoot}%
\usepackage{booktabs}%
\usepackage{algorithm}%
\usepackage{algorithmicx}%
\usepackage{algpseudocode}%
\usepackage{listings}%
\usepackage{array}
\usepackage{ragged2e}
\usepackage{tabularray}
\usepackage{rotating}
\usepackage{pdflscape}
\usepackage{longtable}
\usepackage{tabularray}
\usepackage[T1]{fontenc}

\usepackage{soul}
\usepackage{color}

\begin{document}

\begin{frontmatter}



\title{KACQ-DCNN: Uncertainty-Aware Interpretable Kolmogorov--Arnold Classical--Quantum Dual-Channel Neural Network for Heart Disease Detection}

\author[1,2]{Md Abrar Jahin}
\ead{jahin@usc.edu, abrar.jahin.2652@gmail.com}
\author[3]{Md. Akmol Masud}
\ead{akmolmasud5@gmail.com}
\author[4]{M. F. Mridha\corref{corauthor}}
\ead{firoz.mridha@aiub.edu}
\author[5]{Zeyar Aung}
\ead{zeyar.aung@ku.ac.ae}
\author[6]{Nilanjan Dey}
\ead{nilanjan.dey@tint.edu.in}

\affiliation[1]{organization={Thomas Lord Department of Computer Science},
            addressline={University of Southern California}, 
            city={Los Angeles},
            state={CA},
            postcode={90089}, 
            country={USA}}
\affiliation[2]{organization={Physics and Biology Unit},
            addressline={Okinawa Institute of Science and Technology Graduate University (OIST)}, 
            city={Okinawa},
            postcode={904-0412}, 
            country={Japan}}
\affiliation[3]{organization={Institute of Information Technology},
            addressline={Jahangirnagar University}, 
            city={Savar, Dhaka},
            postcode={1342},
            country={Bangladesh}}
\affiliation[4]{organization={Department of Computer Science},
            addressline={American International University-Bangladesh}, 
            city={Dhaka},
            postcode={1229},
            country={Bangladesh}}
\affiliation[5]{organization={Center for Secure Cyber-Physical Systems (C2PS) and Department of Computer Science},
            addressline={Khalifa University}, 
            city={Abu Dhabi},
            postcode={127788},
            country={UAE}}
\affiliation[6]{organization={Department of Computer Science and Engineering},
            addressline={Techno International New Town}, 
            city={Kolkata},
            postcode={700156},
            state={West Bengal},
            country={India}}

\cortext[corauthor]{Corresponding author}

\begin{abstract}
\textit{Background:} Heart failure remains a critical global health issue, contributing significantly to cardiovascular disease burden and accounting for approximately 17.8 million annual deaths worldwide. Traditional diagnostic approaches face substantial limitations in early detection and intervention planning.\\
\textit{Problem:} Classical machine learning models struggle with complex, high-dimensional data, class imbalances, poor categorical feature representations, and lack interpretability due to their `black box' nature. While quantum machine learning shows potential, existing hybrid models have yet to capitalize on quantum advantages for cardiovascular diagnostics fully.\\
\textit{Solution:} We propose the \ul{K}olmogorov--\ul{A}rnold \ul{C}lassical--\ul{Q}uantum \ul{D}ual-\ul{C}hannel \ul{N}eural \ul{N}etwork (KACQ-DCNN), a novel hybrid architecture that integrates KAN components with quantum circuits, enabling univariate learnable activation functions that significantly improve function approximation with reduced complexity. Our comprehensive evaluation demonstrates that the KACQ-DCNN 4-qubit 1-layered model outperforms 37 benchmark models with 92.03\% accuracy and 94.77\% ROC-AUC score. Ablation studies confirm a synergistic effect of classical-quantum components with KAN, improving accuracy by approximately 2\% compared to MLP variants.\\
\textit{Benefits:} KACQ-DCNN improves heart disease detection accuracy while providing interpretable insights through model-agnostic explainability techniques like Local Interpretable Model-agnostic Explanations (LIME) and Shapley Additive exPlanations (SHAP). Additionally, conformal prediction techniques deliver robust uncertainty quantification, advancing the field toward more reliable, transparent, and clinically trustworthy cardiovascular diagnostic systems that facilitate timely and effective interventions.
\end{abstract}



\end{frontmatter}


\section{Introduction}\label{sec:Introduction}
Heart failure (HF) is a severe health condition in which the heart cannot pump sufficient blood to meet the body's needs, contributing significantly to global morbidity and mortality~\citep{rahman_enhancing_2024}. The increase in HF cases is largely attributed to modern dietary habits, sedentary lifestyles, and increased work-related stress~\citep{babu_revolutionizing_2024,srinivasan_active_2023}. According to the World Health Organization (WHO), cardiovascular disease (CVD) accounts for $\approx17.8$ million deaths annually, with coronary artery disease (CAD) being the most common~\citep{srinivasan_active_2023}. CAD is caused by blockages in the coronary arteries that impede blood flow, leading to severe complications such as heart attack and vascular rupture. This condition affects $\approx2\%$ of the global population and consumes a similar percentage of healthcare budgets~\citep{babu_revolutionizing_2024}, creating a substantial financial burden while reducing individual productivity. This underscores the critical need for preventive and diagnostic strategies to reduce CAD-related mortality.

Traditional heart disease diagnostics rely on assessing risk factors such as age, sex, family history, lifestyle choices, and symptom analysis~\citep{rao_attgru-hmsi_2024}. However, this manual approach often fails to capture the hidden parameters and is computationally expensive~\citep{li_heart_2020,babu_revolutionizing_2024}. Although angiography remains the gold standard for diagnosing CAD~\citep{babu_revolutionizing_2024,rao_attgru-hmsi_2024}, its high cost and potential side effects limit its widespread use. Additionally, image-based diagnostic methods are impractical for large-scale screening because of their high cost. In response, researchers are increasingly focusing on developing noninvasive, cost-effective, and rapid automated diagnostic systems for early CAD detection via machine learning (ML) algorithms. Recent advances in artificial intelligence (AI) and ML have significantly improved diagnostic accuracy and accessibility by leveraging large datasets to uncover nonlinear and hierarchical patterns that traditional clinical methods often overlook. These technologies offer a more effective means of analyzing complex medical data and addressing the uncertainty and lack of interpretability that can lead to misdiagnoses or missed cases in conventional approaches.

Current HF prediction models utilizing classical ML (CML)~\citep{rahman_enhancing_2024,praveena_rachel_kamala_predictive_2023,das_comprehensive_2023,rani_decision_2021,saboor_method_2022,sharma_heart_2020,singh_heart_2020}, deep learning (DL)~\citep{degroat_discovering_2024,rao_attgru-hmsi_2024,nannapaneni_hybrid_2023,shrivastava_hcbilstm_2023}, hybrid~\citep{kavitha_heart_2021,mir_novel_2024,mohan_effective_2019}, and quantum ML (QML)~\citep{babu_revolutionizing_2024,dunjko_quantum-enhanced_2016,wan_quantum_2017,rebentrost_quantum_2014,schetakis_review_2022,yoshioka_hunting_2024,kumar_heart_2021,kavitha_quantum_2023} techniques face several significant research gaps that limit their effectiveness, particularly when applied to short, imbalanced, and tabular datasets characterized by a mix of categorical and numerical features~\citep{jahin_qamplifynet_2023}. The integration of classical--quantum ML (CQML) techniques into heart disease prediction remains largely unexplored despite their potential to increase predictive performance through improved optimization and demonstrate computational capabilities even on short and imbalanced datasets~\citep{jahin_qamplifynet_2023,abbas_power_2021,beer_training_2020}. Similarly, while the latest Kolmogorov--Arnold networks (KANs)~\citep{yu_kan_2024,liu_kan_2024,koenig_kan-odes_2024,jamali_kan_2024} offer theoretical advantages over MLP-based models and CNNs—such as faster convergence using fewer parameters, learnable univariate functions on edges, optimization capability of activation functions, and smoother approximations with B-splines—they have yet to be fully implemented on practical datasets, limiting their real-world applications such as heart failure detection and investigating its interdisciplinary combinations with fields like QML.

Many ML-DL models operate as ``black boxes,'' making it challenging for healthcare professionals to understand the rationale behind predictions. This lack of transparency can hinder trust in these systems, especially when critical clinical decisions are at stake~\citep{hossain_machine_2024,baghdadi_advanced_2023}. Traditional models often fail to quantify the uncertainty in predictions, which is crucial for clinical decision-making. The integration of uncertainty-aware methods within classical--quantum frameworks could enhance the reliability of predictions; however, this has not been adequately explored. Heart disease datasets often suffer from class imbalance, where instances of heart disease are significantly fewer than those of non-disease cases. Traditional models may not adequately address this issue, leading to biased predictions that favor the majority class~\citep{al-alshaikh_comprehensive_2024,baghdadi_advanced_2023}. Additionally, these models frequently struggle with overfitting, especially when trained on small datasets, leading to poor generalizability to unseen data~\citep{babu_revolutionizing_2024,subramani_cardiovascular_2023}. Furthermore, handling categorical features can be inadequate because traditional encoding methods may not effectively capture the underlying relationships. Traditional statistical methods often assume linear relationships between variables, which may not reflect the complex, nonlinear interactions present in cardiovascular health data~\citep{hossain_machine_2024,subramani_cardiovascular_2023}. Moreover, the significant computational demands and extended training durations of DL models can hinder efforts to enhance the accuracy of heart sound classification~\citep{rahman_enhancing_2024}.

We present the ``\ul{K}olmogorov--\ul{A}rnold \ul{C}lassical--\ul{Q}uantum \ul{D}ual-\ul{C}hannel \ul{N}eural \ul{N}etwork (KACQ-DCNN),'' a hybrid model that combines the unique strengths of the KAN, CQML, explainable AI (XAI) methods such as Shapley Additive exPlanations (SHAP) and Local Interpretable Model-agnostic Explanations (LIME), and uncertainty quantification through conformal prediction techniques. KACQ-DCNN addresses multiple challenges in ML by replacing MLP layers with KAN layers, particularly in managing short, imbalanced, high-categorical, and high-dimensional datasets, offering a novel approach to solving real-world problems such as heart disease prediction. At the core of this innovation is the KAN-enhanced CQML model, which uses learnable activation functions on the edges, allowing it to approximate continuous functions with fewer parameters and reduced complexity. Its spline-based edge functions offer smoother and interpretable mappings, enabling faster convergence and improved generalization~\citep{liu_kan_2024}. KAN excels in handling high-dimensional, nonlinearly separable data and provides the optimization facility of the activation function, unlike traditional MLPs and CNNs~\citep{koenig_kan-odes_2024,jamali_kan_2024}. Furthermore, KACQ-DCNN leverages the exponential computational power of quantum techniques, allowing for the faster exploration of vast solution spaces and the ability to capture detailed correlations and patterns. The quantum components significantly boost scalability, making the model adept at handling small and large datasets while retaining computational efficiency~\citep{jahin_qamplifynet_2023}. We selected BiLSTMKANnet as channel-1 and QDenseKANnet as channel-2 after thorough benchmarking, as both models demonstrated superior performance compared to all other evaluated options. Our choice of a dual-channel neural network (DCNN) was driven by its ability to process diverse types of input data in parallel, leveraging distinct feature extraction capabilities for each channel. This approach allows us to combine the individual predictions from each channel to produce final probabilities, effectively enhancing overall performance without compromising efficiency. This architecture enabled us to effectively capture both local and global contexts, which previously worked well with sentiment analysis tasks~\citep{zhang_sentiment_2023}, where understanding nuanced interactions is crucial. Furthermore, it minimizes the risk of overfitting and offers robustness against noise and variability in real-world applications, allowing the model to adapt to dynamic data streams while maintaining accuracy~\citep{ye_dual-channel_2024,zheng_dual-channel_2024}. In addition, incorporating XAI tools such as SHAP and LIME enhances the model’s transparency, offering interpretable insights into predictions, which is especially valuable in fields such as healthcare~\citep{nilanjan}. To ensure reliability, KACQ-DCNN integrates uncertainty quantification through conformal prediction, provides robust confidence intervals, and addresses ambiguous or high-risk classification scenarios.

The novel contributions of this research are as follows:
\begin{enumerate}
\item We developed the KACQ-DCNN framework, which innovatively integrates classical and quantum elements by replacing MLPs with KANs to increase performance. To our knowledge, this represents our initial effort to integrate KANs with a classical--quantum neural network architecture, advancing state-of-the-art ML methodologies.
\item Our study uniquely merges and generalizes data from five major databases with sophisticated preprocessing steps, enhancing model robustness and broadening applicability across diverse populations.
\item We benchmarked KACQ-DCNN against 37 different models, including CML, quantum neural networks (QNNs), hybrid models, and variants of KACQ-DCNN, demonstrating superior performance across various metrics and establishing a new standard in the field.
\item This research executes comprehensive ablation tests on KACQ-DCNN, along with 10-fold cross-validated two-tailed paired t-tests against nine other top-performing models with a corrected 99.44\% confidence level, providing robust statistical evidence of its effectiveness and reliability.
\item We present comparative analyses between KACQ-DCNN and state-of-the-art models in the heart disease prediction literature, highlighting our advancements and contributions to the domain.
\item This study incorporated XAI techniques, including SHAP and LIME, to provide transparency and interpretability of the model's predictions.
\item Finally, we included uncertainty quantification through conformal prediction techniques, ensuring that our model outputs are reliable and trustworthy, followed by estimating $CO_{2}$ emissions in our model experiments.
\end{enumerate}

The remainder of this paper is organized as follows: ``\hyperref[lit_rev]{Related literature}'' reviews related work in heart disease detection using CML and QML techniques; ``\hyperref[sec:methods]{Methods}'' section details the design of the experiments, the proposed KACQ-DCNN framework, the architecture of the benchmarked models, and the training procedure; ``\hyperref[sec:results]{Results and discussion}'' section presents and discusses the experimental results, including performance comparisons with benchmarked and existing models, ablation studies, statistical tests, interpretability analysis, the quantification of uncertainties via conformal prediction, and $CO_{2}$ emissions in our experiments; and ``\hyperref[sec:conclusion]{Conclusion}'' section concludes with future research directions.

\section{Related literature}
\label{lit_rev}
\subsection{Supervised ML and DL Models for Heart Disease Prediction}
Various supervised ML algorithms have been utilized to diagnose heart disease. Recent studies have conducted detailed comparative analyses of the effectiveness of these prediction models for heart disease across widely used datasets. Kamala et al.~\citep{praveena_rachel_kamala_predictive_2023} explored the performance of models such as support vector machines (SVM), Random Forest (RF), and K-nearest neighbors (KNN) on heart disease datasets. Among these, RF consistently outperformed the others across several metrics, including sensitivity, accuracy, and specificity.
Similarly, Das and Sinha~\citep{das_comprehensive_2023} conducted a comprehensive analysis of multiple algorithms, including SVM, Naive Bayes (NB), decision trees (DT), and logistic regression (LR). Their results indicated that SVM and a voting classifier achieved the highest classification accuracy, suggesting that ensemble methods could significantly benefit early diagnosis. Shrivastava et al.~\citep{shrivastava_hcbilstm_2023} developed a hybrid model integrating convolutional neural networks (CNN) and bidirectional long short-term memory (BiLSTM) networks to address issues such as missing data and data imbalance. Their model achieved a 96.66\% accuracy on the Cleveland UCI dataset, illustrating the power of combining DL techniques with ML models to enhance performance. Arooj et al.~\citep{arooj_deep_2022} demonstrated the utility of a deep CNN (DCNN) in heart disease prediction, achieving high accuracy, recall, and F1 scores, with a validation accuracy of 91.7\%. These results highlight the ability of CNNs to outperform traditional ML models when large datasets are available. Nannapaneni et al.~\citep{nannapaneni_hybrid_2023} explored a hybrid model that combined CNNs, LSTM networks, and gated recurrent units (GRU), achieving 93.3\% accuracy, further demonstrating the potential of DL for heart disease detection. However, the complexity of these models can lead to higher computational costs and longer training times, presenting challenges for their clinical deployment. Sahoo et al.~\citep{kumar_sahoo_machine_2022} evaluated a range of ML algorithms, including LR, KNN, SVM, NB, DT, RF, and XGBoost. Their study found that RF achieved the highest accuracy of 90.16\%. Despite this, the study highlighted that no single model performs best across all datasets, and the choice of the optimal model often depends on the specific characteristics of the dataset, such as the size, class balance, and feature distribution. 

Rao et al.~\citep{rao_computational_2021} emphasized that the dataset context is crucial when selecting a model because different models excel under various conditions. This underscores the need to consider model selection criteria carefully, including the trade-off between accuracy and interpretability. Singh and Kumar~\citep{singh_heart_2020} applied a KNN classifier (k=3) to the Cleveland dataset and achieved 87.00\% accuracy, whereas Kavita et al.~\citep{kavitha_heart_2021} employed a hybrid decision tree-RF model, improving the accuracy to 88.00\%. Mohan et al.~\citep{mohan_effective_2019} introduced the HRFLM model, which balances recall and precision and attains an accuracy of 88.40\%. Sharma et al.~\citep{sharma_heart_2020} also achieved a remarkable recall (99.00\%) via RF, reinforcing the consistent performance of this algorithm. Saboor et al.~\citep{saboor_method_2022} reported a 96.72\% accuracy with an SVM classifier, further demonstrating the robustness of CMLs on the Cleveland dataset.

\begin{table*}[!ht]
\caption{Comparative summary of recent studies implementing supervised ML and DL in heart disease prediction}
\label{tab:classical-dl-table}
\resizebox{\textwidth}{!}{%
\begin{tabular}{@{}lllll@{}}
\toprule[1.5pt]
\textbf{Study} & \textbf{Method(s)} & \textbf{Dataset} & \textbf{Accuracy} & \textbf{Highlights} \\ 
\midrule[1pt]
\cite{mir_novel_2024} & RF + Adaboost & Heart Statlog, UCI & Statlog: 99.48\%, UCI: 93.90\% & Outperformed previous studies \\
\cite{praveena_rachel_kamala_predictive_2023} & SVM, RF, KNN & Heart Disease DB & RF: 82.11\% & RF performed best in sensitivity and accuracy \\
\cite{das_comprehensive_2023} & SVM, NB, DT, LR, Voting & Heart Disease DB & SVM/Voting: Best & Ensemble learning was beneficial \\
\cite{shrivastava_hcbilstm_2023} & CNN + BiLSTM (Hybrid) & Cleveland & 96.66\% & Robust to missing data \& imbalance \\
\cite{arooj_deep_2022} & DCNN & Heart Disease DB & 91.70\% & Strong CNN performance \\
\cite{nannapaneni_hybrid_2023} & CNN, LSTM, GRU (Hybrid) & Cleveland & 93.30\% & High performance from hybrid DL \\
\cite{kumar_sahoo_machine_2022} & LR, KNN, SVM, NB, DT, RF, XGBoost & Cleveland & RF: 90.16\% & RF was overall best \\
\cite{singh_heart_2020} & KNN (k=3) & Cleveland & 87\% & Moderate accuracy with simple method \\
\cite{kavitha_heart_2021} & Hybrid DT + RF & Cleveland & 88\% & Hybrid model improved results \\
\cite{mohan_effective_2019} & HRFLM & Cleveland & 88.40\% & Balanced precision and recall \\
\cite{sharma_heart_2020} & SVM, RF, DT, NB & Cleveland & – & RF had better recall \\
\cite{saboor_method_2022} & SVM & Cleveland & 96.72\% & High accuracy via SVM \\
\cite{li_heart_2020} & FCMIM + Classifier & Cleveland & 92\% & Boosted recall with feature selection \\
\cite{doppala_hybrid_2023} & Genetic Algorithm + RBF & Cleveland & 94\% & Achieved high accuracy with optimization \\
\cite{rani_decision_2021} & Random Forest & Cleveland & 86.6\% & Balanced metrics \\
\cite{shah_heart_2020} & KNN (k=7) & Cleveland & 90.78\% & Tuned K produced high accuracy \\
\bottomrule[1.5pt]
\end{tabular}%
}
\end{table*}

\subsection{QML in Medical Prediction}
QML is an emerging field that combines quantum computing with traditional ML techniques, offering the potential to enhance computational power and improve diagnostic accuracy. Although still in its early stages, QML has shown potential for medical applications. Magallanes et al.~\citep{ovalle-magallanes_hybrid_2022} demonstrated the superiority of a hybrid classical--quantum model over CMLs in detecting stenosis from X-ray coronary angiography, significantly improving accuracy, recall, and F1 scores. Gupta et al.~\citep{gupta_comparative_2022} compared QML techniques with DL models for diabetes prognosis using the PIMA Indian Diabetes dataset. Although the DL model achieved higher accuracy (95\%), the QML model still performed well, showing promise in handling large datasets. Maheshwari et al.~\citep{maheshwari_machine_2020} explored the computational efficiency of quantum algorithms compared with classical algorithms in diabetes classification and reported that quantum approaches improved computational speed by up to 55 times. This suggests that although QML may not yet consistently outperform traditional models in terms of accuracy, its ability to handle vast datasets efficiently positions it as a valuable tool for medical prediction. Kumar et al.~\citep{kumar_heart_2021} compared traditional ML with QML algorithms and reported that the QML algorithm achieved an accuracy of 89\% in heart failure detection, with notable improvements in F1-score and recall. These findings suggest that although QML is still being developed, its advantages in handling large, complex datasets and improving computational efficiency are promising. QML has also gained traction; Kavitha and Kaulgud~\citep{kavitha_quantum_2023} presented a quantum K-means algorithm that attained 96.40\% accuracy, marking significant progress in applying quantum computing to heart disease diagnostics. While current applications of QML are limited, the field is rapidly evolving, and future advancements may allow quantum models to surpass classical methods entirely.

\begin{table*}[!ht]
\caption{Comparative summary of recent studies implementing QML in heart and medical prediction tasks}
\label{tab:qml-table}
\resizebox{\textwidth}{!}{%
\begin{tabular}{@{}lllll@{}}
\toprule[1.5pt]
\textbf{Study} & \textbf{Method(s)} & \textbf{Dataset} & \textbf{Accuracy} & \textbf{Highlights} \\ 
\midrule[1pt]
\cite{ovalle-magallanes_hybrid_2022} & CQML (Quantum-Classical Hybrid) & Coronary Angiography & 91.80\% & Outperformed CML baselines \\
\cite{gupta_comparative_2022} & QML vs. DL & PIMA Diabetes & QML: 86\% & QML promising for large data \\
\cite{maheshwari_machine_2020} & QML vs. CML & Diabetes & QML: 69\% & QML was 55x faster \\
\cite{kumar_heart_2021} & QML vs. Classical ML & Heart Failure & QML: 89\% & Better F1 and Recall scores \\
\cite{babu_revolutionizing_2024} & QML vs. CML & Cleveland & QuEML: 93\% & Outperformed CML benchmarks \\
\cite{kavitha_quantum_2023} & Quantum K-means & Cleveland & 96.40\% & Major step for QML diagnostics \\
\cite{contreras_kan-eeg_2024} & KAN-EEG & EEG Seizure Data & – & Generalizable and reduced overfitting \\
\cite{li_u-kan_2024} & Seg. U-KAN (KAN + U-Net) & Medical Imaging & – & Accurate and computationally efficient \\
\bottomrule[1.5pt]
\end{tabular}%
}
\end{table*}

\subsection{KANs in Medical AI}
KANs represent the latest innovative approach to neural network architectures, offering enhanced accuracy with fewer parameters, dynamic adaptability, and architectural flexibility in medical diagnostics. Li et al.~\citep{li_u-kan_2024} developed the U-KAN model, a hybrid approach that combines KANs with U-Net for medical image segmentation tasks. This model reduces computational costs and improves segmentation performance, making it highly suitable for applications such as MRI analysis. In addition, KANs have been used for seizure detection based on EEG data. Contreras et al.~\citep{contreras_kan-eeg_2024} developed the KAN-EEG model, which demonstrated superior generalizability across datasets from different regions, reducing overfitting and improving diagnostic accuracy. The flexibility and adaptability of KANs in medical diagnostics make them valuable, particularly for tasks requiring high interpretability and computational efficiency.

ML has proven to be a valuable tool for heart disease prediction, with many models showing high accuracy and reliability. Although there are multiple databases related to heart disease, most studies have been conducted via the Cleveland database, and a few have used Statlog, demonstrating the need for more diverse and aggregated datasets. Conventional neural networks typically demand a higher number of parameters and extensive computational resources, posing a significant challenge in environments with limited resources. Traditional ML algorithms such as RF and SVM have demonstrated robustness in clinical settings, while hybrid and DL models offer enhanced accuracy for complex datasets. However, previous research focused only on a single type of dataset at a time, which might contain potential bias, lack of generalizability, or overfitting because of being short, highly categorical, and imbalanced. While previous studies explored QML and CQML in heart disease prediction~\citep{babu_revolutionizing_2024,kavitha_quantum_2023,kumar_heart_2021}, showing their promise for improving computational efficiency and diagnostic performance, they have yet to outperform classical methods consistently. KANs, as substitutes for MLP, offer a novel approach to improving the accuracy with less computational complexity and flexibility of neural network architectures in medical diagnostics, which remains greatly unexplored. Integrating KANs with CQML techniques will likely lead to more precise, efficient, and interpretable models for heart disease prediction.

\section{Methods}\label{sec:methods}
\begin{figure*}[!ht]
    \centering
    \includegraphics[width=0.75\linewidth]{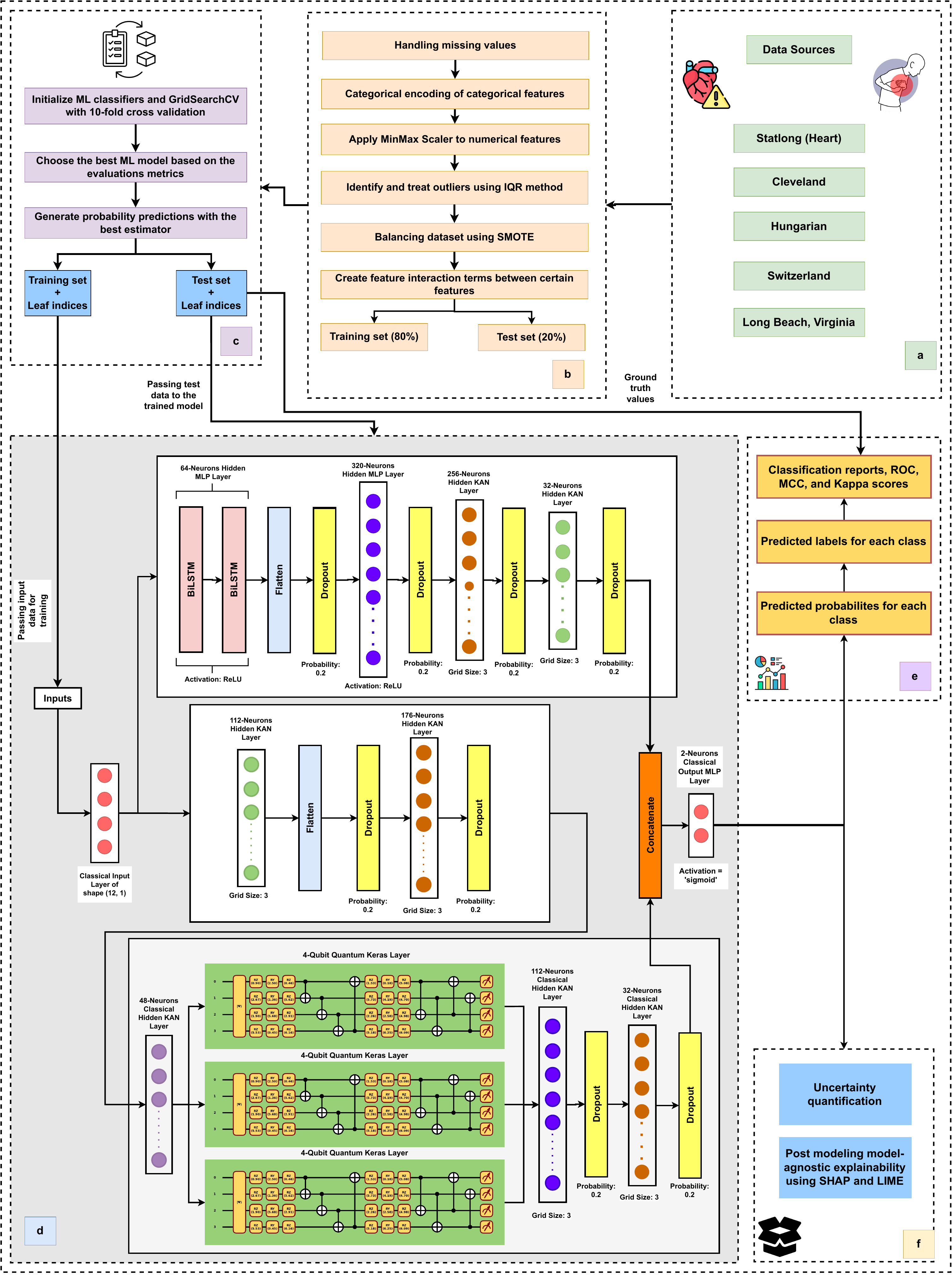}
    \caption{Methodological framework for developing and evaluating the KACQ-DCNN model. (a) \textit{Data Acquisition:} Five heart disease datasets are collected. (b) \textit{Preprocessing \& EDA:} Datasets are merged and subjected to cleaning, encoding, scaling, outlier treatment, SMOTE balancing, and feature interaction creation, followed by an 80/20 train/test split. (c) \textit{Classical Feature Engineering:} The best CML model (identified via GridSearchCV) generates leaf indices from the training data, which are then concatenated with original features for both the train and test sets. (d) \textit{KACQ-DCNN Implementation:} The augmented data is input to the dual-channel KACQ-DCNN (Channel 1: BiLSTMKANnet; Channel 2: QDenseKANnet with quantum layers). Outputs are concatenated and fed to a final classification layer. (e) \textit{Performance Evaluation:} Model predictions are assessed using various metrics (classification reports, ROC-AUC, MCC, Kappa) against ground truth. (f) \textit{Uncertainty \& Interpretability Analysis:} Conformal prediction is used for uncertainty quantification, and SHAP/LIME are employed for model-agnostic explainability.}
    \label{fig:methodology}
\end{figure*}

Figure~\ref{fig:methodology} presents the comprehensive methodology for developing and evaluating our proposed KACQ-DCNN framework. The process initiates with data acquisition, where multiple heart disease datasets from diverse sources are integrated (see Section \ref{sec3.1}). These datasets then undergo extensive preprocessing and exploratory data analysis, including handling missing values, categorical encoding, normalization, outlier detection and management, data balancing through SMOTE, generation of feature interaction terms, and dataset splitting into training and test sets (detailed in Section \ref{sec3.3}). In the classical feature engineering phase, various classical machine learning (CML) classifiers are initialized with GridSearchCV and evaluated via cross-validation to identify the optimal model based on multiple performance metrics. The selected model not only predicts labels on the test set but also outputs class probabilities and a detailed classification report. Additionally, model-derived information, such as leaf indices or probability scores, is combined with the original features to form an enriched feature set for both training and testing data. At the heart of our framework lies the KACQ-DCNN architecture (explained in Section 3.8), a hybrid dual-channel model integrating both classical (BiLSTMKANnet) and dressed quantum (QDenseKANnet) components. These engineered features are processed in parallel through each channel, with their outputs concatenated and passed to a final dense layer followed by a 2-neuron sigmoid-activated output layer for binary classification. The subsequent evaluation phase computes comprehensive classification metrics, predicted labels, and class probabilities (Section \ref{sec4.1.1}). Finally, to ensure clinical interpretability and reliability, a post-hoc analysis phase is conducted, involving uncertainty quantification through conformal prediction (Section \ref{sec4.3}) and model explainability using techniques such as SHAP and LIME (Section \ref{sec4.2}).

\subsection{Dataset description}\label{sec3.1}
The dataset utilized in our research is a consolidated collection derived from five well-known cardiovascular disease datasets with 1190 observations: Statlog (Heart), Cleveland, Hungarian, Switzerland, and Long Beach (Virginia). These datasets contain observations from distinct geographical locations, with 270, 303, 294, 123, and 200 entries, respectively. After eliminating 272 duplicate entries, the final dataset comprised 918 unique observations. This compilation provided a substantial sample size, making it one of the most extensive datasets available for heart disease prediction.

\begin{figure*}[!ht]
    \centering
    \includegraphics[width=0.9\linewidth]{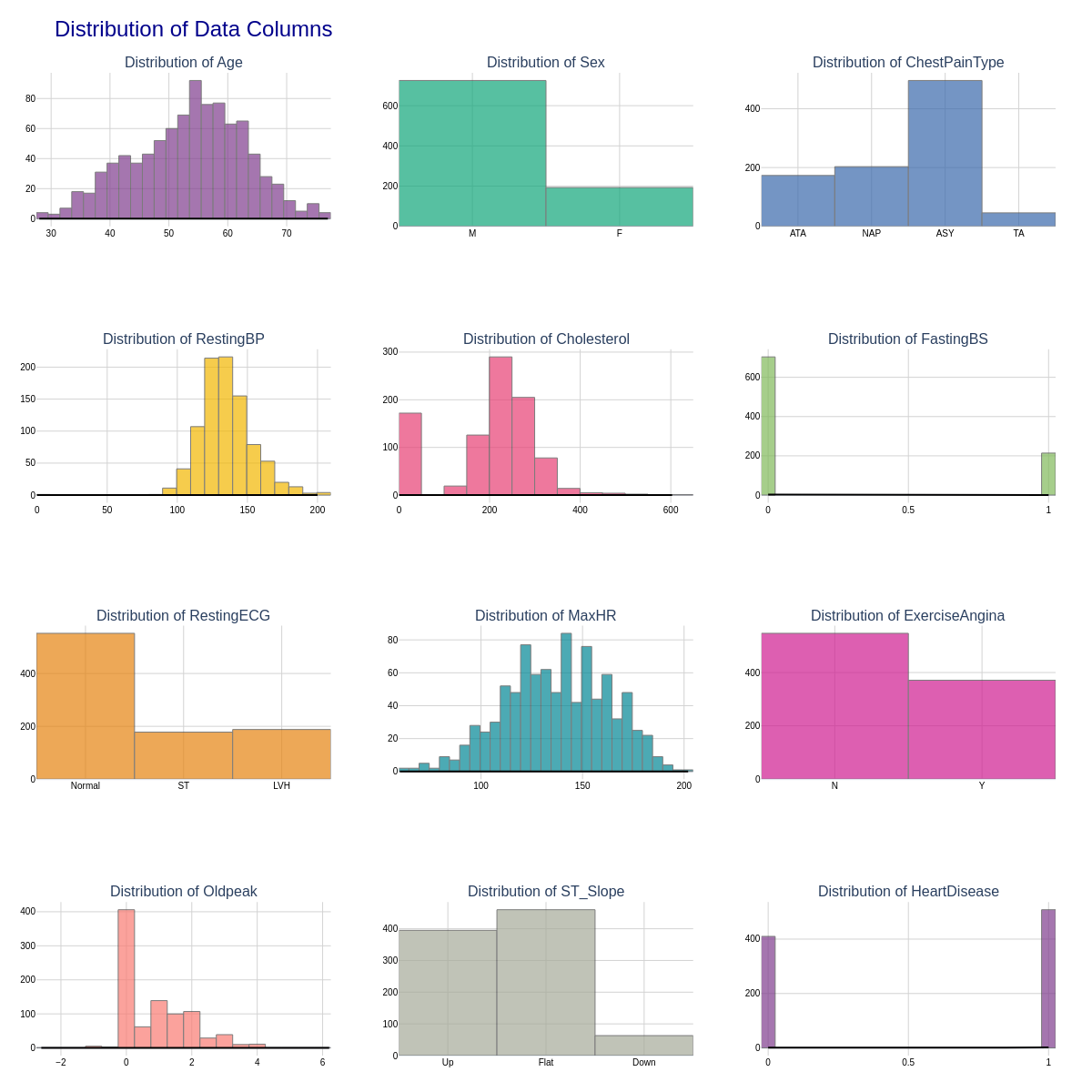}
    \caption{Distribution of cardiovascular health indicators used in our dataset.}
    \label{fig:distributions}
\end{figure*}

The dataset encompasses various features that are instrumental in diagnosing cardiac conditions, as detailed in Table \ref{tab:dataset}. It includes demographic attributes, such as sex and age, which are recognized as critical risk factors for cardiovascular disease. Clinical attributes include resting blood pressure, serum cholesterol levels, and maximum heart rate, each of which plays a pivotal role in assessing heart function. Additionally, the dataset provides detailed information about various types of chest pain, ranging from typical angina to asymptomatic conditions, further assisting in accurately diagnosing heart disease. The distributions of twelve essential variables in our dataset are shown in Figure \ref{fig:distributions}. These variables include demographics (age, sex), clinical measurements (blood pressure, cholesterol, and blood sugar), cardiac function indicators (ECG results and heart rate), and diagnostic markers (ST segment characteristics).

\begin{table}[!ht]
\centering
\caption{Summary of the dataset's key attributes}
\label{tab:dataset}
\begin{tabular}{ m{3cm}  m{12cm} }
\toprule[1.5pt]
\textbf{Attribute} & \textbf{Description} \\ \midrule[1pt]
$Age$ & Age of the patient in years \\ 
$Sex$ & Sex of the patient (M: Male, F: Female) \\ 
$ChestPainType$ & Chest pain type (TA: Typical Angina, ATA: Atypical Angina, NAP: Non-Anginal Pain, ASY: Asymptomatic) \\ 
$RestingBP$ & Resting blood pressure in mm Hg \\ 
$Cholesterol$ & Serum cholesterol in mm/dl \\ 
$FastingBS$ & Fasting blood sugar (1: if FastingBS $>$ 120 mg/dl, 0: otherwise) \\ 
$RestingECG$ & Resting electrocardiogram results (Normal: Normal, ST: ST-T wave abnormality, LVH: Left ventricular hypertrophy) \\ 
$MaxHR$ & Maximum heart rate achieved (numeric value between 60 and 202) \\ 
$ExerciseAngina$ & Exercise-induced angina (Y: Yes, N: No) \\ 
$Oldpeak$ & Oldpeak = ST (numeric value measured in depression) \\ 
$ST\_Slope$ & Slope of the peak exercise ST segment (Up: upsloping, Flat: flat, Down: downsloping) \\ 
$HeartDisease$ & Target class (1: Heart disease, 0: Normal) \\ \bottomrule[1.5pt]
\end{tabular}
\end{table}

Clinical indicators, such as the results from resting electrocardiograms (ECGs), offer valuable insights into heart health, potentially revealing underlying issues such as metabolic disorders or fasting blood sugar irregularities. The dataset also includes an important measure known as ``Oldpeak,'' which quantifies ST-segment depression, a critical indicator of myocardial ischemia. Exercise-induced angina, another key variable, is a strong predictor of heart disease, especially under physical stress conditions. Furthermore, the ST-segment slope during peak exercise provides additional diagnostic information regarding the heart's response to stress, classified as upslope, flat, or downslope. 

\begin{figure*}[!ht]
\centering
\begin{subfigure}{0.47\linewidth} \centering
    \includegraphics[width=\linewidth]{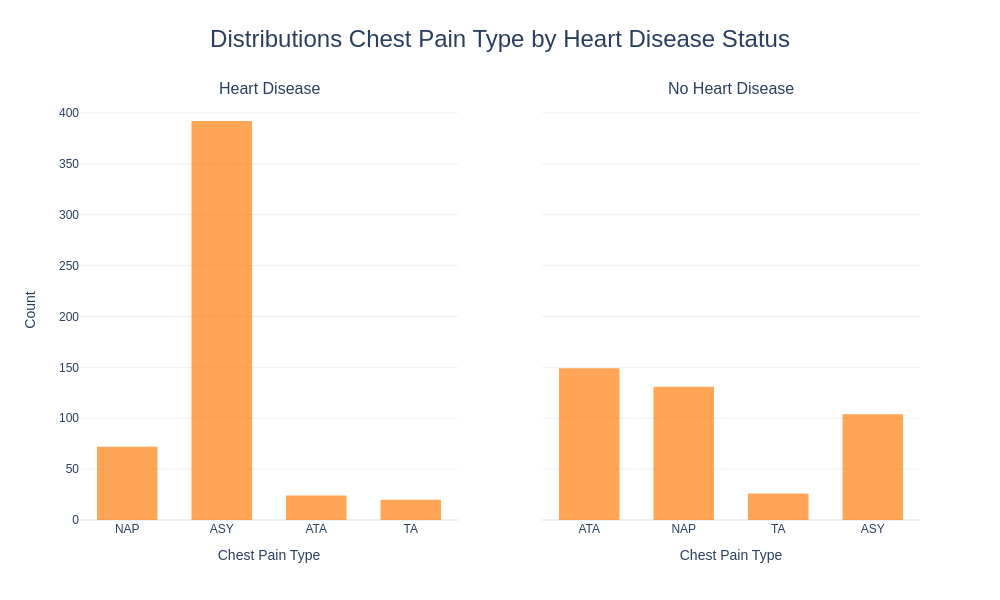}
    \caption{The frequency of different chest pain types (NAP, ASY, ATA, and TA) is compared between individuals with and without heart disease.}
    \label{fig:chest_pain_type}
\end{subfigure}
\hfill
\begin{subfigure}{0.47\linewidth} \centering
    \includegraphics[width=\linewidth]{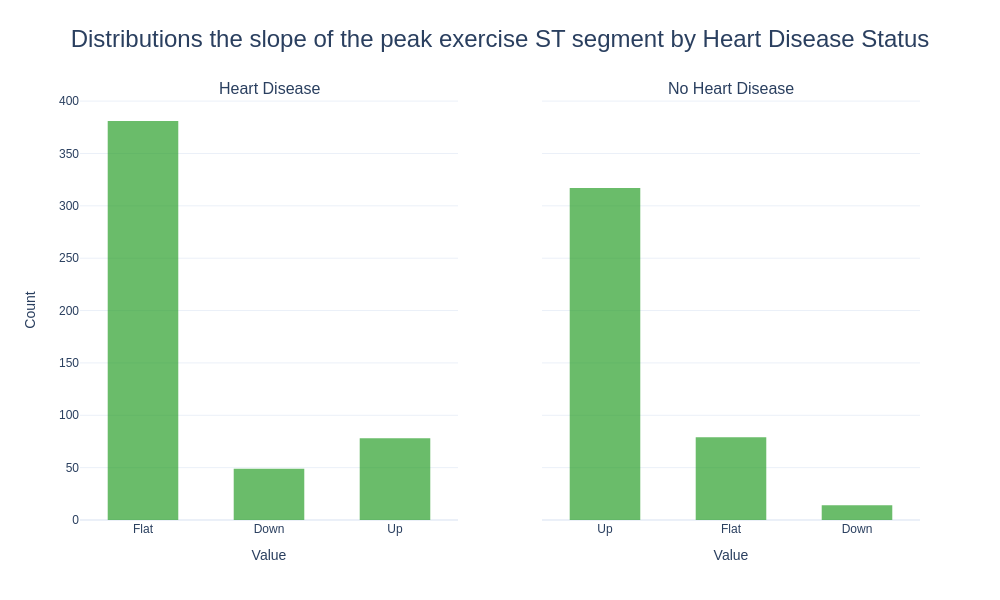}
   \caption{The distribution of ST segment slopes (up, flat, and down) during peak exercise is compared between individuals with and without heart disease.}
    \label{fig:peak_excersize}
\end{subfigure}
\caption{Comparison of chest pain types and ST segment slopes between individuals with heart disease and individuals without heart disease highlighted key differences in ASY, ATA, and `Flat' slope prevalence.}
\label{fig:ST_segment}
\end{figure*}

\begin{figure*}[!ht]
    \centering
    \includegraphics[width=0.8\linewidth]{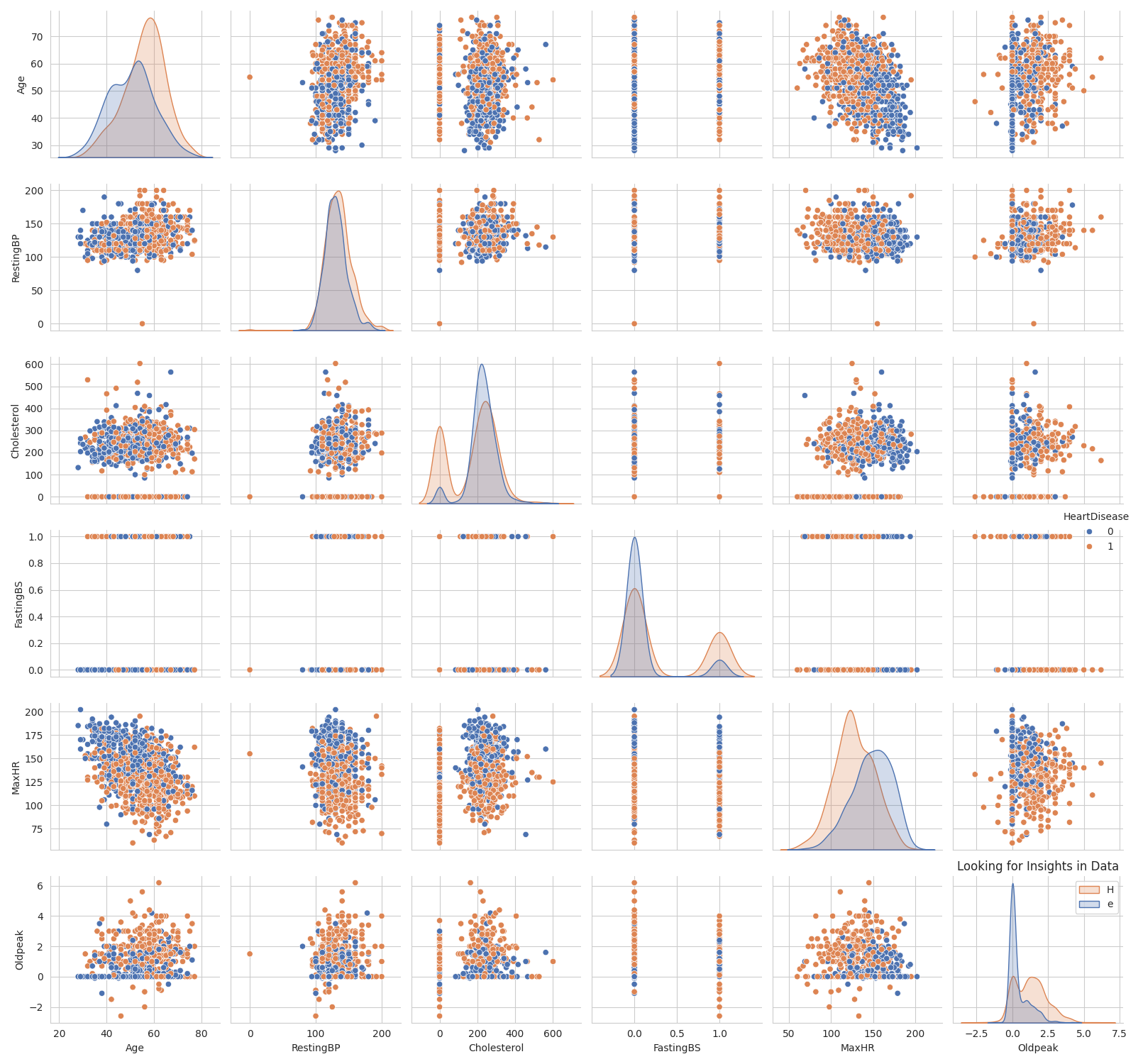}
    \caption{Multidimensional analysis of cardiovascular risk factors and their interrelationships. This matrix of scatter plots and density distributions provides a comprehensive view of key variables associated with heart disease. Each plot compares two variables, with points colored according to heart disease status (orange for the present, blue for absent).}
    \label{fig:pairplot}
\end{figure*}

Notably, asymptomatic (ASY) is most common in heart disease patients, whereas atypical angina (ATA) predominates in non-heart disease subjects, as shown in Figure \ref{fig:chest_pain_type}. Similarly, Figure \ref{fig:peak_excersize} shows a higher prevalence of `Flat' slopes observed in heart disease cases, suggesting potential diagnostic value. Figure \ref{fig:pairplot} displays noteworthy patterns, including $Age$ vs. $RestingBP$ showing a positive correlation, $Cholesterol$ levels displaying a bimodal distribution in heart disease patients, $MaxHR$ negatively correlated with age, and $Oldpeak$ values generally greater in heart disease cases. This figure identifies complex, multivariable relationships that may inform risk assessment and diagnostic strategies for heart disease.

\subsection{Experimental setup}
In our study, we utilized a powerful computational system to evaluate the performance of various models. It is crucial to select appropriate hardware and software to ensure the efficient processing of data, model training, and evaluation. Table \ref{tab:resources} lists the details of the computational resources used in these experiments.

\begin{table}[!ht]
\centering
\caption{Specifications of computational resources}
\label{tab:resources}
\footnotesize
\begin{tabular}{@{}l>{\raggedright\arraybackslash}p{0.8\linewidth}}
\toprule[1.5pt]
\textbf{Resource} & \textbf{Specification}   \\ 
\midrule[1pt]
CPU                & Intel(R) Xeon(R) with x86 architecture, 2 GHz clock speed, 4 virtual cores, 18 GB RAM                     \\ 
GPU                & Dual NVIDIA T4, each with 2560 CUDA cores and 16 GB memory                                                 \\ 
Memory             & 32 GB DDR4                                                                                                  \\ 
Python version     & 3.10.12                                                                                                    \\ 
Libraries/Packages & numpy, pandas, matplotlib, plotly, seaborn, scikit-learn, tensorflow, keras, urllib, zipfile, scipy, crepes, mapie, qiskit, tfkan, pennylane-qiskit, pylatexenc, pennylane, silence-tensorflow, shap, lime \\ 
\bottomrule[1.5pt]
\end{tabular}
\end{table}

\subsection{Data preprocessing}
Our dataset consisted of both categorical and numerical features. Figure \ref{fig:correlation_matrix} illustrates the Pearson correlation matrix revealing notable correlations among features in our dataset, including positive associations with heart disease and $ExerciseAngina$ (0.49) and negative correlations with $MaxHR$ (-0.40) and $ChestPainType$ (-0.39), highlighting key relationships in heart disease diagnosis. The preprocessing steps applied to this dataset were as follows:

\begin{figure*}[!ht]
    \centering
    \includegraphics[width=0.7\linewidth]{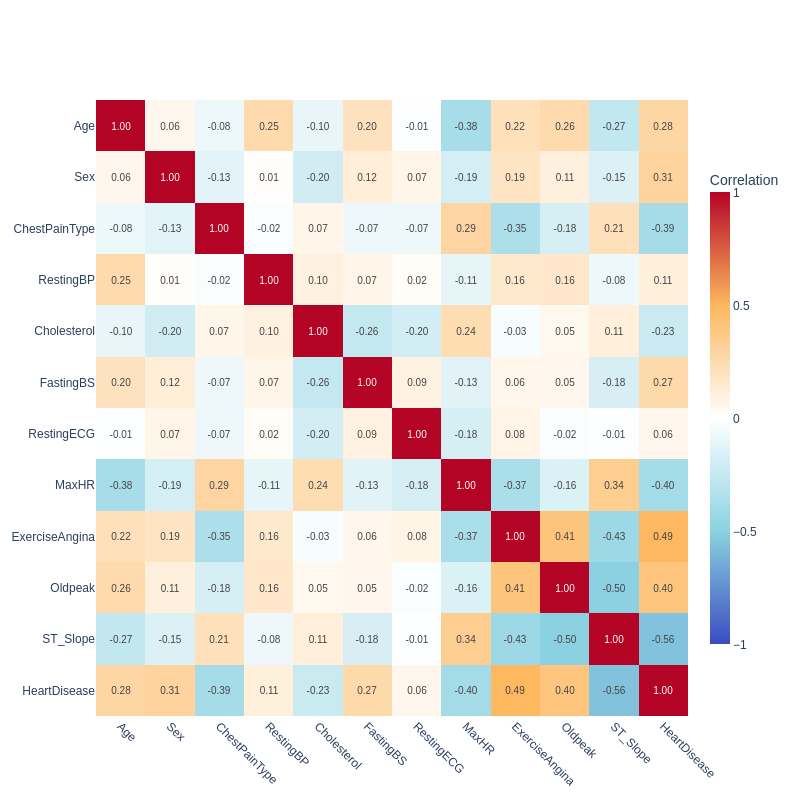}
    \caption{Correlation matrix of cardiovascular risk factors and diagnostic indicators. The heatmap displays pairwise Pearson correlation coefficients between 13 variables related to heart disease. Color intensity and value represent correlation strength and direction. This analysis helps identify multicollinearity and features strongly correlated with the target variable during exploratory data analysis.}
    \label{fig:correlation_matrix}
\end{figure*}

The dataset was checked for missing values, particularly for features such as \textit{Cholesterol}, \textit{RestingBP}, and \textit{MaxHR}. If missing values were present, they were credited with the median value for numerical columns and the mode for categorical columns. This ensures that the dataset remains complete and that no valuable data are discarded.

The \textit{Sex} feature, categorical with two levels (`M' for male and `F' for female), was label-encoded as 0 and 1, respectively. This encoding allows the model to interpret gender differences numerically. The \textit{ChestPainType} feature, consisting of categories (`TA,' `ATA,' `NAP,' `ASY'), was one-hot encoded to create separate binary columns for each type. This avoids ordinal assumptions and allows the model to independently understand each chest pain type's presence. The \textit{RestingECG} feature, containing values (`Normal,' `ST,' and `LVH'), was also one-hot encoded. This ensures the model can differentiate between the ECG results without implying any ordinal relationship. The binary \textit{ExerciseAngina} feature (`Y' for yes, `N' for no) was label-encoded to 1 and 0, respectively. The \textit{ST\_Slope} feature, which has three categories (`Up,' `Flat,' `Down'), was one-hot encoded to allow the model to capture the distinct impact of different slope types on heart disease prediction.

Numerical features such as \textit{Age}, \textit{RestingBP}, \textit{Cholesterol}, \textit{MaxHR}, and \textit{Oldpeak} were scaled via min-max scaling to transform the data into a range between 0 and 1. This scaling method ensures that all numerical features contribute equally to the learning process of the model, particularly in algorithms where the scale of the data can influence performance.

Outliers in numerical features were identified and addressed via the interquartile range (IQR) method. Any values outside $1.5 \times IQR$ were treated as potential outliers and were either capped or transformed to reduce their impact on the model's performance.

The dataset was checked for class distribution because the target variable \textit{HeartDisease} could be imbalanced. When an imbalance was found, methods such as the synthetic minority oversampling technique (SMOTE) or changing class weights during model training were considered to ensure the model performed well in both classes. Some features, such as \textit{Age} and \textit{MaxHR} or \textit{ChestPainType} and \textit{ExerciseAngina}, have interaction terms that were used to find any nonlinear relationships that could help the model work better. Finally, the dataset was split into training and testing sets via an 80:20 split to ensure the performance of the models could be evaluated on unseen data. A stratified sampling method was used to maintain the proportion of the target classes in both the training and testing sets.

\subsection{Classical conventional ML models}
We predicted heart illness via several conventional ML models employing standard and ensemble techniques. The benchmarked algorithmic models were DT-based algorithms, ensemble methods, probabilistic classifiers, and linear models. CatBoost, LR, KNN, RF, histogram-based gradient boosting, LightGBM, extreme gradient boosting (XGBoost), gradient boosting, AdaBoost, bagging classifier, and DT were applied. Much attention has been paid to Bernoulli NB, MLP, linear discriminant analysis (LDA), and quadratic discriminant analysis (QDA) classifiers. The relative baseline consisted of a dummy classifier. 

Every model assessed in the test set used GridSearchCV to optimize the model by searching for the best hyperparameters, specifically focusing on the number of iterations. Stratified 10-fold cross-validation was used to evaluate overfitting and model generalizability. This method guarantees a constant means of assessing the model's performance on fresh data by ensuring that every fold has the same target variable distribution. After identifying CatBoost as the best-performing CML model, it was used to predict the labels on the test set. These probabilities were subsequently combined with the original feature sets via column stacking, resulting in augmented feature matrices for training and test sets. This step aims to enrich the feature space by incorporating the learned representations of the model.

\subsection{Quantum neural networks}
In our quantum circuit implementation, multi-scale entanglement renormalization ansatz (MERA), matrix product states (MPS), and tree tensor networks (TTN) were used to process the input features through amplitude embedding (AE), which normalized and padded the data for quantum computation. The circuit output was then evaluated using the expected value of the Pauli-Z operator. Several optimizers, such as Nesterov momentum and Adam, were used to find the best values for the MERA, MPS, and TTN parameters so that the square loss between predicted and actual labels was as small as possible. The performance of the MPS network was evaluated by varying the number of layers from 1 to 4. The training process involved splitting the dataset into training and validation sets, optimizing the weights via the selected optimizer, and computing the classifier accuracy.

\subsubsection{MERA}
MERA is a sophisticated tensor network structure designed to efficiently capture and represent the entanglement properties of quantum many-body systems, particularly those with scale-invariant or critical behavior. Unlike more superficial tensor network structures such as MPS or TTN, MERA incorporates disentangling operations known as isometries, which systematically reduce correlations at different scales. This hierarchical approach allows MERA to represent quantum states with long-range entanglement, thus making it a powerful tool for studying critical phenomena in quantum systems. We employed the MERA structure to model the complex correlations between the dataset features. Our implementation uses a variational quantum classifier (VQC) framework, where the MERA layers act as entangling blocks of the quantum circuit. Specifically, the quantum circuit comprises a series of $CNOT$ gates and $RY$ rotations controlled by trainable parameters. In quantum computing, the CNOT gate is a 2-qubit gate that flips the state of the target qubit if the control qubit is in the \( |1\rangle \) state and can be represented as a $4\times4$ unitary matrix:
\begin{equation}\label{eq:cnot}
\text{CNOT} = 
\begin{pmatrix}
1 & 0 & 0 & 0 \\
0 & 1 & 0 & 0 \\
0 & 0 & 0 & 1 \\
0 & 0 & 1 & 0 \\
\end{pmatrix}
\end{equation}

The input data were encoded into the quantum circuit via AE, ensuring that the input features were normalized and appropriately padded. Mathematically, the MERA circuit can be expressed as a composition of local unitary operations \( U \) and isometries \( W \), which act on the quantum state \( |\psi\rangle \). For a single layer of MERA, the wavefunction can be expressed as:
\begin{equation}
|\psi_{\text{MERA}}\rangle = U^{(1)} W^{(1)} U^{(2)} W^{(2)} \cdots U^{(L)} W^{(L)} |\psi_0\rangle,
\end{equation}
where \( L \) represents the number of layers, each \( W^{(i)} \) disentangles the correlations at scale \( i \), and \( U^{(i)} \) applies local unitary operations.

\subsubsection{MPS}
MPS represents a powerful and versatile tensor network approach for modeling quantum many-body systems. In the MPS framework, the quantum state of a one-dimensional system is expressed as the product of matrices, each associated with a site in the chain. This representation is beneficial for capturing local correlations and is foundational for developing algorithms such as density matrix renormalization group (DMRG).

In our research, we applied the MPS structure via a quantum VQC, which involves a quantum circuit in which the MPS layers are represented by a series of quantum gates, including CNOT gates and RY rotations. Each layer of the MPS circuit was parameterized with trainable weights, enabling the model to adapt and optimize its parameters for classification tasks. Mathematically, an MPS representation for a quantum state \( |\psi\rangle \) in a one-dimensional chain with \( n \) sites can be expressed as:
\begin{equation}
|\psi_{\text{MPS}}\rangle = \sum_{i_1, i_2, \ldots, i_n} A^{i_1} A^{i_2} \cdots A^{i_n} |i_1, i_2, \ldots, i_n\rangle,
\end{equation}
where \( A^{i_k} \) is the matrix associated with the \( k \)-th site, and \( |i_k\rangle \) is the basis state. Each matrix \( A^{i_k} \) typically has fixed dimensions, making the representation efficient for computation and storage.

\subsubsection{TTN}
TTN~\citep{schollwock_density-matrix_2005,evenbly_entanglement_2009} provides a hierarchical structure for modeling complex quantum states and is particularly effective for capturing correlations at multiple scales. Unlike simpler tensor network models such as MPS, TTN organizes tensors in a tree-like structure. This provides a more detailed picture of how subsystems interact and become entangled.

In our study, we applied the TTN architecture to a quantum VQC. The TTN approach uses a tree-based arrangement of tensors, where each tree level captures correlations on a different scale. This hierarchical structure is beneficial for modeling quantum states with multi-scale entanglement, which is advantageous when dealing with complex datasets. TTN can be described mathematically as follows: a quantum state \( |\psi\rangle \) is represented by a network of tensors organized in a tree structure. For a given tensor network with \( L \) levels, the state can be expressed as:
\begin{equation}
|\psi_{\text{TTN}}\rangle = \sum_{i_1, i_2, \ldots, i_n} \prod_{l=1}^L T^{(l)}_{i_1 i_2 \cdots i_{n_l}} |i_1, i_2, \ldots, i_n\rangle,
\end{equation}
where \( T^{(l)} \) represents the tensor at level \( l \) of the tree, and \( n_l \) denotes the number of indices associated with the tensor at that level.

\subsection{Hybrid models}
In our study, we developed, tuned, and evaluated three distinct hybrid models to assess the performance of BiLSTM networks, KAN-based CNNs, and KAN-based feed-forward dense networks.

\subsubsection{BiLSTMKANnet}
The BiLSTMKANnet model integrates BiLSTM networks with KAN to process sequential data while ensuring model interpretability. The model begins by accepting input tensors, which are run through two stacked BiLSTM layers, with 64 units each, and ReLU activation, which picks up both forward and backward temporal dependencies in the sequence. Standard equations govern the internal mechanisms of each LSTM cell: the forget gate \(f_t = \sigma(W_f \cdot [h_{t-1}, X_t] + b_f)\), input gate \(i_t = \sigma(W_i \cdot [h_{t-1}, X_t] + b_i)\), cell state update \(C_t = f_t \cdot C_{t-1} + i_t \cdot \tanh(W_C \cdot [h_{t-1}, X_t] + b_C)\), and output gate \(o_t = \sigma(W_o \cdot [h_{t-1}, X_t] + b_o)\), where \(h_t = o_t \cdot \tanh(C_t)\) represents the hidden state. The outputs of the forward and backward passes in BiLSTM are concatenated, yielding a robust representation, \(H_t = \text{concat}(\overrightarrow{h_t}, \overleftarrow{h_t})\), of the input sequence from both temporal directions.

Following the BiLSTM layers, the sequence data are flattened into a dense vector representation, transitioning the model from sequential to dense processing. This flattened output was then passed through a dense layer with 320 neurons, activated by ReLU \(D_1 = \text{ReLU}(W_1 \cdot H_{\text{flattened}} + b_1)\), followed by a dropout layer with a rate of 20\% to mitigate overfitting, \(D_1^{\text{dropout}} = \text{Dropout}(D_1, \, \text{rate}=0.20)\). The model’s inclusion of DenseKAN layers is crucial, as the KAN structure leverages Kolmogorov--Arnold representations to leverage its dynamic activation functions, learnable univariate functions, and architectural flexibility. The DenseKAN layers perform functional approximation via a grid of basis functions, effectively mapping complex relationships within the data. The first DenseKAN layer has 256 units and a grid size of 3, defined mathematically as \(K_1 = \sum_{i=1}^{n} \sum_{j=1}^{m} a_{ij} \cdot \phi_i(x_j)\), followed by a second DenseKAN layer with 32 units, each interspersed with dropout layers to ensure regularization, \(K_2^{\text{dropout}} = \text{Dropout}(K_2, \, \text{rate}=0.20)\).

This architecture allows the BiLSTMKANnet model to harness the sequence modeling strengths of BiLSTMs while integrating the interpretability and representational power of KANs. The result is a model capable of delivering accurate predictions with an inherent ability to explain the decision-making process, making it particularly suitable for applications requiring high performance and transparency.

\subsubsection{QCKANnet}
The QCKANnet model combines classical Convolution KAN (Conv-KAN) layers, feed-forward dense KAN layers, and quantum computing elements to enhance data processing capabilities and dynamic adaptability. The model starts with input tensors, which pass through two Conv1DKAN layers. These layers apply convolutional operations with 96 filters, kernel sizes of 3 and 2, and strides of 2 while utilizing KAN structures with a grid size of 3 to capture complex patterns. Unlike standard convolutions, Conv1DKAN replaces static kernels with dynamic, learnable functions based on the Kolmogorov--Arnold theorem. This allows kernel values to adjust during training, improving adaptation to different data regions. Conv-KAN layers achieve higher accuracy with fewer parameters, reducing computational demands and increasing efficiency, ideal for resource-limited settings. The output is then flattened and transformed through a DenseKAN layer with \((2^{n_{\text{qubits}}}) \times 3\) units, where \(n_{\text{qubits}}\) is set to 4, effectively expanding the feature space before applying a series of quantum operations.

The model employs three quantum layers, \( \text{qlayer}_1, \text{qlayer}_2, \text{qlayer}_3 \), each defined by a quantum node that uses AE and strongly entangling layers (SEL) with parameterized quantum gates. We evaluated the model with 1, 2, 3, and 4 layers of quantum operations to determine the optimal configuration for the task. The quantum layers are input parts of the dense output split into three segments, and quantum operations are applied to these segments, leveraging the quantum circuit's ability to model complex interactions. These segments are then concatenated and passed through additional DenseKAN layers, featuring 256 and 32 units with a grid size of 3, respectively, followed by dropout layers for regularization. The final layer is a dense layer with a sigmoid activation function that outputs data dimensions suitable for binary classification tasks. By integrating Conv-KAN operations, feed-forward KAN layers, and quantum components, QCKANnet offers a sophisticated architecture that leverages the strength of each domain to achieve enhanced performance and interpretability.

\subsubsection{QDenseKANnet}
The QDenseKANnet model is a hybrid classical--quantum neural network architecture that integrates quantum computing principles with KAN for advanced data processing. The model begins by accepting an input tensor and processing it through a series of fully connected DenseKAN layers, which utilize dynamically adjusted activation functions to capture complex, nonlinear relationships in the data. The first DenseKAN layer has 112 units with a grid size of 3, and its output is flattened and regularized through a dropout layer with a 20\% dropout rate. This was followed by a second DenseKAN layer with 176 units, which was similarly regularized with dropout.

To incorporate quantum computation, the model applies a third DenseKAN layer with a specific dimensionality tailored to match the quantum processing requirements, effectively dressing the quantum circuits with classical preprocessing. This layer's output is split into three parts, each inputting a different quantum layer. Each quantum layer is implemented as a quantum circuit via the PennyLane library, with four qubits and strongly entangling gates. The quantum circuits are QNodes, where the inputs are embedded into the quantum state via AE, and the quantum operations are applied via SEL. The output of each quantum circuit consists of the expected values of Pauli-Z measurements on each qubit, which represent the quantum features extracted from the data. We evaluated the model with 1, 2, 3, and 4 layers of quantum operations to determine the optimal configuration for the task.

These quantum features were then concatenated and passed through additional DenseKAN layers to further process and refine the data. The final DenseKAN layers have 112 and 32 units, each interspersed with dropout layers to maintain regularization. The DenseKAN layers are defined by the equation \( K(x) = \sum_{i=1}^{n} \sum_{j=1}^{m} a_{ij} \cdot \phi_i(x_j) \), where \(a_{ij}\) represents the coefficients and \(\phi_i(x_j)\) represents the basis functions defined by the grid size. The quantum layers, implemented as quantum neural networks, are defined by the quantum node function \( qnode(\text{weights}, \text{inputs}) \), which applies quantum operations and returns the expectation values of Pauli-Z operators. By combining the interpretability and functional approximation capabilities of KANs with the computational power of quantum circuits, the QCKANnet model is designed to solve complex problems requiring high predictive accuracy, dynamic adaptability, and flexibility for handling high-dimensional data.

\subsection{Proposed KACQ-DCNN}\label{sec3.3}
As channel-1 and channel-2, we utilized BiLSTMKANnet (4-qubit 1-layered) and QDenseKANnet (4-qubit 1-layered), respectively, based on their comparatively better performance among the hybrid models. KACQ-DCNN consists of 3,045,891 parameters ($\approx14$ MB in size), of which 3,023,329 parameters are trainable and are updated during the training process to optimize the model's performance. The remaining 22,562 parameters are non-trainable and typically remain fixed, such as the parameters from pre-trained layers or fixed embeddings used within the model. First, we discuss how the quantum circuit architecture was developed and how it performs necessary computations. Second, we discuss the architecture of KANs implemented in KACQ-DCNN. Finally, we discuss the integration of BiLSTMKANnet and dressed quantum circuit QDenseKANnet.

\subsubsection{Architecture of quantum circuit}
\begin{figure*}[!ht]
    \centering
    \includegraphics[width=1\linewidth]{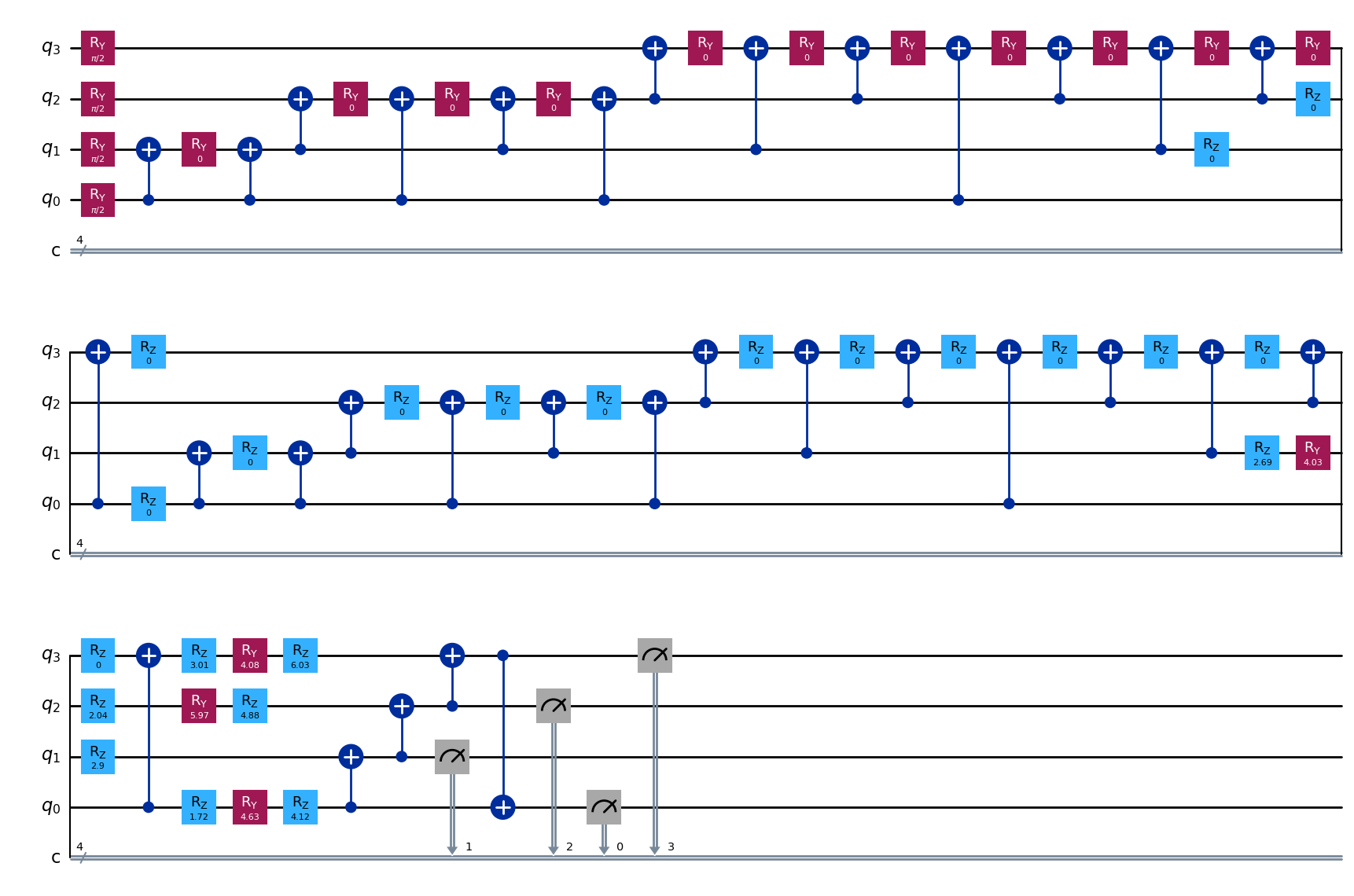}
    \caption{The KACQ-DCNN quantum circuit is structured with 4 qubits labeled `0', `1', `2', and `3'. The architecture incorporates a single variational layer applied across all 4 qubits. The initial quantum state is prepared via AE, where classical information is encoded into the quantum system, as represented by the circuit's first set of blue lines. Each qubit is then rotated into a superposition state via a series of $R_Y$ gates, with rotation angles of $\frac{\pi}{2}$ determined by the parameters of the network. These rotations allow the qubits to explore a broader range of quantum states. Following this, parameterized U3 gates (Equation \ref{eq:u3}) are applied to each qubit, with the gate parameters optimized during the training phase to improve the network's performance. After the first layer, entanglement is introduced through a series of CNOT gates, creating correlations between the qubits that classical models cannot capture. The circuit then repeats this process with a second layer, where U3 gates are again applied to the qubits. In the final stage, measurements are taken via $Pauli-Z$ operators, represented by the vertical bars on the circuit, which determine the state of each qubit along the Z-axis, providing output values of either +1 or -1 (visualized via Pennylane-Qiskit tools).}
    \label{fig:quantum_cicuit}
\end{figure*}
The quantum circuit architecture (Figure \ref{fig:quantum_cicuit}) employed in our model integrates several critical components designed to effectively encode classical data into quantum states and process them via quantum entanglement. The circuit follows a structured approach, beginning with AE, transitioning through SEL, and culminating in measurement operations. Each component plays a pivotal role in a quantum processing system's overall functionality and efficiency.
\\
\\
\textit{Amplitude Embedding (AE):}
The AE technique transforms classical data into quantum states at the core of the KACQ-DCNN quantum circuit. In this process, a classical data point \( \mathbf{x} \) with \( N \) features is represented as the quantum state \( |\psi_{\mathbf{x}}\rangle \) in an \( n \)-qubit system, where \( N = 2^n \). The quantum state is mathematically expressed as follows:
\begin{equation}
|\psi_{\mathbf{x}}\rangle = \sum_{i=1}^{N} x_i |i\rangle
\end{equation}
Here, \( x_i \) represents the \( i \)-th element of the normalized classical data vector \( \mathbf{x} \), and \( |i\rangle \) denotes the computational basis states. To effectively prepare the quantum state, the classical feature vector \(\mathbf{x}\) is first normalized, ensuring that the sum of the squares of its elements equals 1:
\begin{equation}
\sum_{i} |x_i|^2 = 1
\end{equation}

Following normalization, the AE translates the classical data into a quantum state \( |\psi\rangle \) expressed as a linear combination of basis states \( |i\rangle \):
\begin{equation}
|\psi\rangle = \sum_{i=0}^{N-1} x_i |i\rangle
\end{equation}
where \( x_i \) denotes the normalized values of the classical features corresponding to the basis states. In this circuit, we employ the \( R_y \) rotation gate with an angle of \( \frac{\pi}{2} \) to facilitate the embedding of the classical information into the quantum state. The \( R_y(\theta) \) gate performs a rotation around the Y-axis of the Bloch sphere and is defined mathematically as:
\begin{equation}
R_y(\theta) = \cos\left(\frac{\theta}{2}\right) |0\rangle\langle0| + \sin\left(\frac{\theta}{2}\right) |1\rangle\langle1|
\end{equation}

For our specific case with \(\theta = \frac{\pi}{2}\), the rotation operator simplifies to:
\begin{equation}
R_y\left(\frac{\pi}{2}\right) = \frac{1}{\sqrt{2}} |0\rangle\langle0| + \frac{1}{\sqrt{2}} |1\rangle\langle1|
\end{equation}

This rotation operation transforms the quantum state \( |\psi\rangle \) as follows:
\begin{equation}
|\psi'\rangle = R_y\left(\frac{\pi}{2}\right) |\psi\rangle
\end{equation}

After applying the \( R_y \) gate, the updated quantum state incorporates the classical information encoded in the amplitudes while adjusting the orientation of the state on the Bloch sphere. Consequently, the resulting state can be expressed as:
\begin{equation}
|\psi'\rangle = \frac{1}{\sqrt{2}}(x_0 |0\rangle + x_1 |1\rangle + x_2 |2\rangle)
\end{equation}
The amplitude-embedded quantum state can then be manipulated through additional gates in the strongly entangling ansatz, facilitating entanglement and computation.
\\
\\
\textit{Strongly Entangling Ansatz:}  
Following embedding, the circuit includes a strongly entangling ansatz consisting of alternating sequences of single-qubit rotations and multi-qubit gates (entanglers). These layers facilitate entanglement generation among the qubits, which is crucial for quantum information processing. The architecture employs \( L \) layers, where each layer consists of a set of parameters that govern the rotations and entangling operations:
\begin{equation}
\text{Layer}_l: \text{Rotations} \rightarrow \text{Entanglers}
\end{equation}

The single-qubit rotations implemented in our circuit can be represented via \( R_y \), \( R_z \), and \( U3 \) rotation gates. The U3 gate allows for arbitrary rotation about the Bloch sphere and is defined in Equation \ref{eq:u3}, where the parameters \( \theta, \phi, \) and \( \lambda \) control the rotation angles, enabling the qubit to be prepared in any desired quantum state.
\begin{equation}\label{eq:u3}
U3(\theta, \phi, \lambda) = \begin{pmatrix}
\cos(\theta/2) & -e^{i\lambda} \sin(\theta/2) \\
e^{i\phi} \sin(\theta/2) & e^{i(\phi + \lambda)} \cos(\theta/2)
\end{pmatrix}
\end{equation}
The \( R_y \) rotation gate is defined as:
\begin{equation}\label{eq:ry}
R_y(\theta) = e^{-i\frac{\theta}{2}Y} = \cos\left(\frac{\theta}{2}\right) I - i \sin\left(\frac{\theta}{2}\right) Y = \begin{pmatrix}
\cos\left(\frac{\theta}{2}\right) & -\sin\left(\frac{\theta}{2}\right) \\
\sin\left(\frac{\theta}{2}\right) & \cos\left(\frac{\theta}{2}\right)
\end{pmatrix}
\end{equation}
The \( R_z \) rotation is defined as:
\begin{equation}\label{eq:rz}
R_z(\phi) = e^{-i\frac{\phi}{2}Z} = \begin{pmatrix}
e^{-i\phi/2} & 0 \\
0 & e^{i\phi/2}
\end{pmatrix}
\end{equation}

Multi-qubit entanglers may be implemented as two-qubit CNOT gates (see Equation \ref{eq:cnot}), which act on qubits indexed \( i \) and \( (i+r) \mod M \), where \( r \) is a hyperparameter that defines the range of entanglement. The action of the CNOT gate can be mathematically expressed as:
\begin{equation}
\text{CNOT}_{ij} |x_i, x_j\rangle = |x_i, x_j \oplus x_i\rangle
\end{equation}
where \( x_i \) and \( x_j \) are the states of the control and target qubits, respectively, and where \( \oplus \) denotes addition modulo 2. The layer structure can be mathematically represented as follows:
\begin{equation}
\text{Entangling Layer}(w) = \sum_{j=1}^{M} \left( R_y(\theta_j) \otimes R_z(\phi_j) \otimes \text{Entangler}(q_{i}, q_{(i+r) \mod M}) \right)
\end{equation}
where \( R_y(\theta_j) \) and \( R_z(\phi_j) \) denote the application of the \( R_y \) and \( R_z \) single-qubit rotation gates, parameterized by \( \theta_j \) and \( \phi_j \), respectively. This flexibility allows the architecture to adapt to various configurations and complexities of entangling operations. The optimization of these weights is critical for achieving the desired performance and efficiency in quantum circuits. The choice of rotations and entanglers ultimately influences the ability of the circuit to model complex relationships, such as those found in heart disease prediction datasets.
\\
\\
\textit{Measurement:}  
In the final stage of the KACQ-DCNN quantum circuit, the measurement operation is essential for collapsing the quantum state into a classical output that can inform the diagnosis. This process involves several mathematical steps: application of the Pauli-Z operator, measurement operator, measurement process, and output utilization. Before the measurement, we applied the Pauli-Z operator \( Z \) to the quantum state, which is defined as:
\begin{equation}
Z = \begin{pmatrix}
1 & 0 \\
0 & -1
\end{pmatrix}
\end{equation}
This operator modifies the quantum state by introducing phase shifts to the \( |1\rangle \) state:
\begin{equation}
Z|\psi\rangle = Z\left(\sum_{i} \alpha_i |i\rangle\right) = \sum_{i} \alpha_i Z|i\rangle
\end{equation}
Specifically, for a two-level quantum system, the application of \( Z \) to the basis states is given by: \( Z|0\rangle = |0\rangle \) and \( Z|1\rangle = -|1\rangle \). Thus, applying the Pauli-Z operator results in:
\begin{equation}
Z|\psi\rangle = \alpha_0 |0\rangle - \alpha_1 |1\rangle
\end{equation}
Next, the measurement operator \( M \) is defined based on the desired output. Heart disease prediction can be modeled as a projection operator onto a subspace defined by a binary outcome (e.g., presence or absence of heart disease). Mathematically, this operator can be represented as:
\begin{equation}
M = |0\rangle\langle0| + |1\rangle\langle1|
\end{equation}
Here, \( |0\rangle \) represents the absence of disease, and \( |1\rangle \) represents the presence of disease. The measurement collapses state \( Z|\psi\rangle \) into one of the basis states, yielding probabilities for each outcome. The probability of obtaining the outcome corresponding to \( |j\rangle \) (either \( |0\rangle \) or \( |1\rangle \)) can be calculated via:
\begin{equation}
P(j) = \langle Z\psi | M_j | Z\psi \rangle
\end{equation}
where \( M_j \) is the measurement operator associated with outcome \( |j\rangle \). Specifically, for a binary classification of heart disease prediction:
\begin{equation}
P(0) = \langle Z\psi | M_0 | Z\psi \rangle = |\langle 0| Z|\psi\rangle|^2 = |\alpha_0|^2
\end{equation}
\begin{equation}
P(1) = \langle Z\psi | M_1 | Z\psi \rangle = |\langle 1| Z|\psi\rangle|^2 = |\alpha_1|^2
\end{equation}

This means that the probability of predicting no heart disease is \( |\alpha_0|^2 \), and the probability of predicting heart disease is \( |\alpha_1|^2 \). The probabilities obtained from the measurement are utilized in subsequent classical learning tasks. For example, if the predicted probability of heart disease (presence) exceeds a certain threshold \( \tau \), a diagnosis can be made:
\begin{equation}
\text{Diagnosis} = 
\begin{cases} 
\text{Heart Disease} & \text{if } P(1) > \tau \\ 
\text{No Heart Disease} & \text{if } P(1) \leq \tau 
\end{cases}
\end{equation}

\subsubsection{Architecture of KAN}
KAN is a neural network architecture based on the Kolmogorov--Arnold representation theorem and is designed to decompose complex multivariable functions into more straightforward univariate functions. This decomposition allows KAN to model highly nonlinear relationships more flexibly than traditional MLPs and CNNs. In our implementation within the KACQ-DCNN framework, KAN enhances the dynamic adaptability and flexibility of the neural network through unique architectural elements that reflect the principles of the Kolmogorov--Arnold theorem.

KAN’s architecture departs from conventional MLPs and CNNs by replacing fixed linear weights with learnable univariate functions. These functions, implemented via spline-based techniques, dynamically adjust during training. The KAN consists of multiple layers where nodes in each layer represent the input from the previous layer. The edges between nodes are associated with univariate functions rather than scalar weights, allowing for more granular control over the transformations applied to inputs. A key feature of KAN is that instead of applying fixed activation functions such as ReLU or sigmoid, each weight parameter is replaced by a learnable univariate function. This flexibility allows KAN to dynamically adapt its transformations to better capture complex data patterns. The Kolmogorov--Arnold theorem suggests that any continuous multivariable function \( f(x_1, x_2, \ldots, x_n) \) can be decomposed as follows:
\begin{equation}
f(x_1, x_2, \ldots, x_n) = \sum_{i=1}^{2n+1} g_i \left( \sum_{j=1}^n h_{ij}(x_j) \right)
\end{equation}

This is translated into the network architecture in KAN by applying nested univariate transformations to multivariable inputs. For each input dimension \(x_j\), the network applies a learnable function \(h_{ij}(x_j)\) followed by another univariate transformation \(g_i\) to aggregate the results. In our implementation, the B-spline functions approximate these learnable univariate functions. B-splines are chosen because they provide smooth approximations and are computationally efficient, making them suitable for the dynamic adjustments required during training.

Unlike traditional neural networks that rely on fixed activation functions, KAN dynamically adjusts the activation functions during training. These functions are parameterized via spline functions that allow the model to adapt to the complexity of data. To stabilize training, we implement residual connections via base functions such as the SiLU activation function \( \Phi(x) = \frac{x}{1 + e^{-x}} \), augmented by spline functions. This residual approach ensures smoother gradients during backpropagation, facilitating faster convergence. Instead of optimizing static weights, KAN optimizes the parameters of spline functions. This allows for representing weights as smooth, nonlinear input functions, enhancing the model’s expressiveness and adaptability to various data distributions. Each spline function is initialized to near zero to avoid overpowering the initial training dynamics. The parameters are drawn from a normal distribution \( N(0, \sigma^2) \), with \( \sigma = 0.1 \), and the spline grids are updated dynamically during training to handle large activations that might exceed the predefined range of the spline functions.

We integrated DenseKAN layers in KACQ-DCNN, which correspond to traditional MLP-based dense layers but leverage the KAN framework to adjust weights and activations dynamically via learnable univariate functions. The DenseKAN layer is equivalent to a basic dense (fully connected) layer in traditional neural networks. However, it distinguishes itself by replacing the fixed kernel (weight matrix) with a dynamic, learnable kernel. This layer utilizes learnable univariate functions to represent each weight, enabling the approximation of highly complex functions and enhancing expressiveness in modeling nonlinear relationships within the data. Kernel function \( f \) is updated dynamically based on the Kolmogorov--Arnold theorem, providing increased flexibility. Each weight is represented by univariate functions \( g_i(x) \) and \( h_{ij}(x) \), which adapt during training according to the following equation:
\begin{equation}
y_i = g_i \left( \sum_{j=1}^{n} h_{ij}(x_j) \right)
\end{equation}

\subsubsection{Integrating classical and dressed quantum circuit}
Integrating the BiLSTMKANnet and QDenseKANnet models is designed as a dual-channel architecture that leverages both classical DL and quantum-enhanced computations. The architecture begins by processing the input data separately through two channels: classical BiLSTMKANnet and hybrid quantum-classical QDenseKANnet. Both models accept input tensors of shape \((\text{None}, 12, 1)\), meaning samples with 12 features, each of them represented as a single value, and 1 can be interpreted as the feature dimension, commonly used in neural networks to handle input with multiple features in a consistent shape. However, due to their strengths, the data undergo distinct transformations within each network.

While BiLSTMs are traditionally used for sequential tasks, they are crucial for learning complex relationships within tabular data by capturing both forward and backward dependencies in feature interactions. The output from BiLSTM layers is then passed through DenseKAN layers, which introduce dynamic activation functions that adapt to the data during training. This adaptability enhances the model’s expressiveness, as each weight parameter in the KAN is represented as a univariate function that adjusts to evolving data patterns, ensuring flexibility and accuracy. Moreover, the DenseKAN layers dynamically update their grids based on input activations, allowing them to handle out-of-bounds activation values and refine the model's predictions. 

In the second channel, QDenseKANnet, classical dense layers are combined with quantum operations to enhance feature extraction. DenseKAN layers preprocess the input, encoding classical features before splitting the data into segments passed through quantum circuits. These circuits use AE and SEL to extract features from higher-dimensional spaces, improving the model's ability to capture nonlinear relationships. The term ``dressed quantum circuit'' describes this hybrid model, where classical layers ``dress'' quantum circuits, enhancing their performance while addressing challenges such as noise and qubit limitations. The classical layers refine the input before and after quantum processing, resulting in a more robust and interpretable model for high-dimensional data.

Using a dual-channel architecture for tabular data offers a hybrid approach that capitalizes on the strengths of classical feature modeling and quantum-enhanced feature extraction. The combined output of both channels provides a comprehensive view of tabular data, where the classical channel contributes to model transparency and interpretability, whereas the quantum channel enhances the model’s ability to capture nonlinear interactions between the features. Mathematically, the final output \( Y \) is obtained by concatenating the outputs of the two channels:
\begin{equation}
Y = \sigma\left( W_{\text{concat}} \cdot \left[\text{BiLSTMKANnet}(X), \, \text{QDenseKANnet}(X)\right] + b_{\text{concat}} \right)
\end{equation}
where \( W_{\text{concat}} \) represents the weights of the final fully connected layer, \( b_{\text{concat}} \) is the bias term, and \( \sigma \) is a sigmoid activation function, which is suitable for binary classification of the presence or absence of heart disease.

\subsection{Training and tuning process}
The training and tuning process started by compiling the model via the Adam optimizer, which was set at a learning rate of 0.001, and binary cross-entropy was used as the loss function, whereas accuracy was used as the primary evaluation metric. The training was executed over 100 epochs utilizing a combination of callbacks. First, the best weights of the model are saved on the basis of validation accuracy improvement, ensuring only the optimal model is preserved. A custom callback monitors validation loss and adjusts the learning rate by a factor of 0.5 if there is no improvement after five epochs. Before reducing the learning rate, the model backtracks to the best weights to avoid performance degradation. Finally, the early stopping mechanism stops training if validation loss does not improve after ten epochs while restoring the best weights. This integrated tuning process, alongside validation data, ensures efficient learning by adjusting learning rates and preventing overfitting.

For stratified 10-fold cross-validation, all models went through 10 rounds of experiments, each comprising a split, training, and testing phase. The dataset was divided into 10 equal folds in each round, with 9 folds used for training and 1 fold for testing. This process was repeated 10 times to ensure that each fold was used for testing exactly once. The reported `$\pm$' values in Table \ref{tab:perf_compar} and \ref{tab:ablation} represent the mean and standard deviation of the performance metrics across these 10 rounds, providing a robust measure of both the average performance and the variability for each model.

In our hyperparameter tuning process, we employed the Keras tuner to optimize the parameters of the classical components in BiLSTMKANnet, QCKANnet, and QDenseKANnet models. We used the optimized versions of BiLSTMKANnet and QDenseKANnet for KACQ-DCNN. The hyperband tuner was chosen because of its efficiency in resource allocation across different hyperparameter trials. For BiLSTMKANnet, we initialized the tuner by defining the search space for the LSTM units, dense layer, and KAN layer units. The tuner explored LSTM units in the range of 32--128 in step 32 and a dense layer in the range of 128--512 in step 64.

Similarly, DenseKAN layer units were searched for between 32 and 256. Hyperband was used with a maximum of 10 epochs per trial, and early stopping with patience for 3 epochs was applied to prevent overfitting. This process yielded the best configuration by maximizing the validation accuracy across 50 epochs, and the optimal set of hyperparameters was extracted from one trial. For QCKANnet, the tuner explored filters for Conv1DKAN layers in the range of 32-128 in steps of 32. The Keras tuner used the same settings as in BiLSTMKANnet to maximize validation accuracy while adhering to the early stopping criterion. Finally, QDenseKANnet was used to evaluate the first, second, and fourth DenseKAN layers from 16 to 256 units in step 32. The best configuration was selected on the basis of validation accuracy, and the hyperband facilitated rapid convergence through efficient resource allocation.

\section{Results and Discussion}
\label{sec:results}
\subsection{Model performance}
\subsubsection{Evaluation metrics}\label{sec4.1.1}
In this study, we employed several evaluation metrics to assess the performance of the classification models, including macro average precision (maP), macro average recall (maR), macro average F1-score (maF1), accuracy, ROC-AUC, Matthews correlation coefficient (MCC), and Cohen's kappa. These metrics are based on the counts of true positives (TPs), which are the correctly predicted positive instances; false positives (FPs), which are incorrectly predicted positive instances; false negatives (FNs), which refer to missed positive instances; and true negatives (TNs), which represent the correctly predicted negative instances. This approach provides a comprehensive view of model performance, particularly in scenarios with class imbalance, ensuring that the evaluation accurately captures both positive and negative predictions.
\\
\\
\textbf{Macro average precision (maP):} Precision refers to the ability of the classifier to correctly identify positive instances, i.e., the proportion of TPs among all predicted positives (TP + FP). maP is the average precision score calculated across all classes, treating each class equally irrespective of its size:
\begin{equation}
\text{Precision} = \frac{TP}{TP + FP}
\end{equation}
\begin{equation}
\text{maP} = \frac{1}{N} \sum_{i=1}^{N} \frac{TP_i}{TP_i + FP_i}
\end{equation}
where \( N \) denotes the total number of classes. maP ensures a balanced evaluation across all classes, making it useful for imbalanced datasets.
\\
\\
\textbf{Macro average recall (maR):} Recall measures the model's ability to correctly identify all relevant instances of each class. It is defined as the proportion of true positives among all actual positives (TP + FN):
\begin{equation}
\text{Recall} = \frac{TP}{TP + FN}
\end{equation}
\begin{equation}
\text{maR} = \frac{1}{N} \sum_{i=1}^{N} \frac{TP_i}{TP_i + FN_i}
\end{equation}
maR provides an equal contribution from each class to the final score, ensuring that larger classes do not overshadow smaller classes.
\\
\\
\textbf{Macro average F1-score (maF1):} The F1-score is the harmonic mean of precision and recall. It balances precision and recall, making it useful when the costs of FPs and FNs are high. maF1 averages the F1 scores across all classes:
\begin{equation}
\text{F1-score} = 2 \times \frac{\text{Precision} \times \text{Recall}}{\text{Precision} + \text{Recall}}
\end{equation}
\begin{equation}
\text{maF1} = \frac{1}{N} \sum_{i=1}^{N} 2 \times \frac{\text{Precision}_i \times \text{Recall}_i}{\text{Precision}_i + \text{Recall}_i}
\end{equation}
This score provides a balanced view of precision and recall, which is crucial for imbalanced datasets.
\\
\\
\textbf{Accuracy:} Accuracy measures the proportion of correctly predicted instances (true positives and true negatives) among all the cases:
\begin{equation}
\text{Accuracy} = \frac{TP + TN}{TP + TN + FP + FN}
\end{equation}
Although accuracy is simple and intuitive, it may not always be informative for imbalanced datasets, as it tends to favor the majority class.
\\
\\
\textbf{ROC-AUC:} Receiver operating characteristic — area under the curve (ROC-AUC) evaluates the classifier's ability to distinguish between classes. The ROC curve plots the true positive rate against the false positive rate at various threshold settings. The AUC value, which ranges from 0 to 1, represents the probability that a randomly chosen positive instance is ranked higher than a randomly chosen negative instance.
\\
\\
\textbf{Matthews correlation coefficient (MCC):} MCC considers all four confusion matrix components: TP, TN, FP, and FN. This method is particularly effective for imbalanced datasets:
\begin{equation}
\text{MCC} = \frac{(TP \times TN) - (FP \times FN)}{\sqrt{(TP + FP)(TP + FN)(TN + FP)(TN + FN)}}
\end{equation}
MCC values range from -1 (perfect misclassification) to 1 (perfect classification), with 0 indicating random guessing.
\\
\\
\textbf{Cohen’s kappa:} Cohen’s kappa measures the agreement between predicted and true labels, adjusting for the probability of chance agreement. It is given by:
\begin{equation}
\text{Kappa} = \frac{p_o - p_e}{1 - p_e}
\end{equation}
where \( p_o \) is the observed agreement and \( p_e \) is the expected agreement by chance. Kappa values range from -1 (complete disagreement) to 1 (perfect agreement), with 0 indicating agreement by chance.

These evaluation metrics allow us to assess how well classification models work in many areas, such as accuracy, recall, class imbalance resistance, and agreement between predicted and actual labels.

\begin{table}[!ht]
\caption{Competitive performance benchmarking of the models developed in this research}
\label{tab:perf_compar}
\resizebox{\columnwidth}{!}{%
\begin{tabular}{c l >{\centering\arraybackslash}p{2cm}>{\centering\arraybackslash}p{2cm}>{\centering\arraybackslash}p{0.02\linewidth}cccc c c}
\toprule[1.5pt]
\multirow{2}{*}{\textbf{Model Category}} &
  \multicolumn{1}{c}{\multirow{2}{*}{\textbf{Models}}} &
  \multicolumn{7}{c}{\textbf{Evaluation Metrics}} &
  \multirow{2}{*}{\textbf{Training Time (s) ($\downarrow$)}} &
  \multirow{2}{*}{\textbf{Inference Time (s) ($\downarrow$)}} \\ \cline{3-9}
 &
  \multicolumn{1}{c}{} &
  \multicolumn{1}{c}{\textbf{maP ($\uparrow$)}} &
  \multicolumn{1}{c}{\textbf{maR ($\uparrow$)}} &
  \multicolumn{1}{c}{\textbf{maF1 ($\uparrow$)}} &
  \multicolumn{1}{c}{\textbf{Accuracy ($\uparrow$)}} &
  \multicolumn{1}{c}{\textbf{ROC-AUC ($\uparrow$)}} &
  \multicolumn{1}{c}{\textbf{MCC ($\uparrow$)}} &
  \textbf{Kappa ($\uparrow$)} &
   &
   \\ \midrule[1pt]
\multirow{16}{*}{CML Models} &
  CatBoost &
  \multicolumn{1}{c}{87.18\%$\pm$1.37\%} &
  \multicolumn{1}{c}{90.36\%$\pm$1.10\%} &
  \multicolumn{1}{c}{85.90\%$\pm$1.27\%} &
  \multicolumn{1}{c}{86.43\%$\pm$1.06\%} &
  \multicolumn{1}{c}{91.97\%$\pm$0.49\%} &
  \multicolumn{1}{c}{76.63\%$\pm$1.21\%} &
  76.59\%$\pm$1.15\% &
  1.7556 &
  0.0077 \\  
 &
  multilayer Perceptron &
  \multicolumn{1}{c}{87.26\%$\pm$1.42\%} &
  \multicolumn{1}{c}{85.55\%$\pm$0.84\%} &
  \multicolumn{1}{c}{85.18\%$\pm$1.58\%} &
  \multicolumn{1}{c}{85.94\%$\pm$0.92\%} &
  \multicolumn{1}{c}{90.59\%$\pm$1.42\%} &
  \multicolumn{1}{c}{69.17\%$\pm$0.84\%} &
  69.01\%$\pm$1.23\% &
  0.1014 &
  0.0030 \\  
 &
  Logistic Regression &
  \multicolumn{1}{c}{88.40\%$\pm$0.68\%} &
  \multicolumn{1}{c}{85.36\%$\pm$1.58\%} &
  \multicolumn{1}{c}{85.13\%$\pm$0.53\%} &
  \multicolumn{1}{c}{85.38\%$\pm$1.24\%} &
  \multicolumn{1}{c}{90.31\%$\pm$0.49\%} &
  \multicolumn{1}{c}{66.71\%$\pm$1.46\%} &
  66.69\%$\pm$1.14\% &
  0.1701 &
  0.0230 \\  
 &
  K-Nearest Neighbor &
  \multicolumn{1}{c}{86.17\%$\pm$0.80\%} &
  \multicolumn{1}{c}{87.45\%$\pm$1.08\%} &
  \multicolumn{1}{c}{83.69\%$\pm$0.50\%} &
  \multicolumn{1}{c}{84.33\%$\pm$0.99\%} &
  \multicolumn{1}{c}{89.59\%$\pm$1.36\%} &
  \multicolumn{1}{c}{67.76\%$\pm$1.42\%} &
  67.75\%$\pm$1.26\% &
  0.1071 &
  0.0062 \\  
 &
  Linear Discriminant Analysis &
  \multicolumn{1}{c}{88.38\%$\pm$1.09\%} &
  \multicolumn{1}{c}{83.55\%$\pm$1.15\%} &
  \multicolumn{1}{c}{84.08\%$\pm$0.79\%} &
  \multicolumn{1}{c}{84.30\%$\pm$0.72\%} &
  \multicolumn{1}{c}{90.07\%$\pm$0.51\%} &
  \multicolumn{1}{c}{64.63\%$\pm$1.81\%} &
  64.57\%$\pm$1.67\% &
  0.7587 &
  0.0362 \\  
 &
  Random Forest &
  \multicolumn{1}{c}{86.41\%$\pm$0.57\%} &
  \multicolumn{1}{c}{88.36\%$\pm$1.30\%} &
  \multicolumn{1}{c}{86.07\%$\pm$1.08\%} &
  \multicolumn{1}{c}{84.27\%$\pm$0.94\%} &
  \multicolumn{1}{c}{92.20\%$\pm$1.02\%} &
  \multicolumn{1}{c}{74.35\%$\pm$0.82\%} &
  74.35\%$\pm$0.66\% &
  0.2303 &
  0.0029 \\  
 &
  Quadratic Discriminant Analysis &
  \multicolumn{1}{c}{84.79\%$\pm$0.68\%} &
  \multicolumn{1}{c}{88.36\%$\pm$1.48\%} &
  \multicolumn{1}{c}{83.84\%$\pm$1.26\%} &
  \multicolumn{1}{c}{84.27\%$\pm$0.58\%} &
  \multicolumn{1}{c}{91.06\%$\pm$1.16\%} &
  \multicolumn{1}{c}{62.22\%$\pm$0.91\%} &
  62.22\%$\pm$1.06\% &
  0.8869 &
  0.0021 \\  
 &
  Histogram-based Gradient Boosting &
  \multicolumn{1}{c}{85.77\%$\pm$1.74\%} &
  \multicolumn{1}{c}{86.36\%$\pm$1.16\%} &
  \multicolumn{1}{c}{83.31\%$\pm$0.79\%} &
  \multicolumn{1}{c}{83.80\%$\pm$1.28\%} &
  \multicolumn{1}{c}{89.16\%$\pm$1.25\%} &
  \multicolumn{1}{c}{69.96\%$\pm$1.34\%} &
  69.95\%$\pm$1.21\% &
  13.3002 &
  0.0107 \\  
 &
  LightGBM &
  \multicolumn{1}{c}{85.64\%$\pm$0.60\%} &
  \multicolumn{1}{c}{87.36\%$\pm$1.34\%} &
  \multicolumn{1}{c}{83.05\%$\pm$0.67\%} &
  \multicolumn{1}{c}{83.77\%$\pm$0.85\%} &
  \multicolumn{1}{c}{89.45\%$\pm$0.70\%} &
  \multicolumn{1}{c}{73.40\%$\pm$0.69\%} &
  73.33\%$\pm$1.42\% &
  0.0014 &
  0.0009 \\  
 &
  Extreme Gradient Boosting &
  \multicolumn{1}{c}{85.22\%$\pm$0.88\%} &
  \multicolumn{1}{c}{86.45\%$\pm$0.73\%} &
  \multicolumn{1}{c}{82.64\%$\pm$1.42\%} &
  \multicolumn{1}{c}{83.22\%$\pm$0.73\%} &
  \multicolumn{1}{c}{88.35\%$\pm$1.14\%} &
  \multicolumn{1}{c}{69.17\%$\pm$0.94\%} &
  69.01\%$\pm$1.26\% &
  0.0010 &
  0.0006 \\  
 &
  Gradient Boosting &
  \multicolumn{1}{c}{84.59\%$\pm$1.35\%} &
  \multicolumn{1}{c}{84.73\%$\pm$0.86\%} &
  \multicolumn{1}{c}{81.51\%$\pm$1.27\%} &
  \multicolumn{1}{c}{82.19\%$\pm$1.40\%} &
  \multicolumn{1}{c}{88.33\%$\pm$0.62\%} &
  \multicolumn{1}{c}{77.62\%$\pm$0.80\%} &
  77.62\%$\pm$1.32\% &
  0.0080 &
  0.0008 \\  
 &
  AdaBoost &
  \multicolumn{1}{c}{84.95\%$\pm$0.68\%} &
  \multicolumn{1}{c}{83.55\%$\pm$1.27\%} &
  \multicolumn{1}{c}{81.67\%$\pm$0.76\%} &
  \multicolumn{1}{c}{82.11\%$\pm$0.83\%} &
  \multicolumn{1}{c}{86.90\%$\pm$0.58\%} &
  \multicolumn{1}{c}{68.82\%$\pm$1.11\%} &
  68.82\%$\pm$1.23\% &
  0.0029 &
  0.0007 \\  
 &
  Bagging Classifier &
  \multicolumn{1}{c}{83.70\%$\pm$1.27\%} &
  \multicolumn{1}{c}{85.55\%$\pm$1.39\%} &
  \multicolumn{1}{c}{77.82\%$\pm$0.54\%} &
  \multicolumn{1}{c}{81.55\%$\pm$1.96\%} &
  \multicolumn{1}{c}{89.90\%$\pm$1.55\%} &
  \multicolumn{1}{c}{70.05\%$\pm$1.40\%} &
  70.01\%$\pm$1.02\% &
  0.0014 &
  0.0361 \\  
 &
  Bernoulli Naive Bayes &
  \multicolumn{1}{c}{85.28\%$\pm$1.46\%} &
  \multicolumn{1}{c}{80.45\%$\pm$0.96\%} &
  \multicolumn{1}{c}{79.97\%$\pm$0.88\%} &
  \multicolumn{1}{c}{80.58\%$\pm$1.08\%} &
  \multicolumn{1}{c}{89.51\%$\pm$0.59\%} &
  \multicolumn{1}{c}{57.22\%$\pm$0.85\%} &
  57.04\%$\pm$0.51\% &
  0.0016 &
  0.0009 \\  
 &
  Decision Tree &
  \multicolumn{1}{c}{78.34\%$\pm$2.51\%} &
  \multicolumn{1}{c}{72.82\%$\pm$0.82\%} &
  \multicolumn{1}{c}{71.14\%$\pm$1.68\%} &
  \multicolumn{1}{c}{71.29\%$\pm$1.74\%} &
  \multicolumn{1}{c}{71.85\%$\pm$2.79\%} &
  \multicolumn{1}{c}{49.15\%$\pm$1.10\%} &
  49.13\%$\pm$0.78\% &
  0.1027 &
  0.0125 \\  
 &
  Dummy Classifier &
  \multicolumn{1}{c}{55.44\%$\pm$6.97\%} &
  \multicolumn{1}{c}{100.00\%$\pm$0.61\%} &
  \multicolumn{1}{c}{35.66\%$\pm$5.28\%} &
  \multicolumn{1}{c}{55.44\%$\pm$6.48\%} &
  \multicolumn{1}{c}{50.00\%$\pm$0.47\%} &
  \multicolumn{1}{c}{0.00\%$\pm$0.00\%} &
  0.00\%$\pm$0.00\% &
  0.0299 &
  0.0044 \\ \hline
\multirow{12}{*}{QNN Models} &
  MERA 1-Layered VQC &
  \multicolumn{1}{c}{53.00\%$\pm$0.88\%} &
  \multicolumn{1}{c}{51.00\%$\pm$1.48\%} &
  \multicolumn{1}{c}{44.00\%$\pm$1.24\%} &
  \multicolumn{1}{c}{59.00\%$\pm$0.77\%} &
  \multicolumn{1}{c}{51.00\%$\pm$1.48\%} &
  \multicolumn{1}{c}{3.30\%$\pm$1.73\%} &
  2.16\%$\pm$1.54\% &
  2436.7284 &
  11.88648 \\  
 &
  MERA 2-Layered VQC &
  \multicolumn{1}{c}{20.00\%$\pm$1.32\%} &
  \multicolumn{1}{c}{50.00\%$\pm$3.12\%} &
  \multicolumn{1}{c}{29.00\%$\pm$1.58\%} &
  \multicolumn{1}{c}{41.00\%$\pm$2.20\%} &
  \multicolumn{1}{c}{50.00\%$\pm$1.17\%} &
  \multicolumn{1}{c}{0.00\%$\pm$0.00\%} &
  0.00\%$\pm$0.00\% &
  2377.296 &
  11.88648 \\  
 &
  MERA 3-Layered VQC &
  \multicolumn{1}{c}{30.00\%$\pm$3.67\%} &
  \multicolumn{1}{c}{50.00\%$\pm$2.39\%} &
  \multicolumn{1}{c}{37.00\%$\pm$0.18\%} &
  \multicolumn{1}{c}{59.00\%$\pm$1.03\%} &
  \multicolumn{1}{c}{50.00\%$\pm$1.37\%} &
  \multicolumn{1}{c}{0.00\%$\pm$0.00\%} &
  0.00\%$\pm$0.00\% &
  2555.5932 &
  11.88648 \\  
 &
  MERA 4-Layered VQC &
  \multicolumn{1}{c}{20.00\%$\pm$1.46\%} &
  \multicolumn{1}{c}{50.00\%$\pm$1.78\%} &
  \multicolumn{1}{c}{29.00\%$\pm$0.98\%} &
  \multicolumn{1}{c}{41.00\%$\pm$2.65\%} &
  \multicolumn{1}{c}{50.00\%$\pm$0.75\%} &
  \multicolumn{1}{c}{0.00\%$\pm$0.00\%} &
  0.00\%$\pm$0.00\% &
  2436.7284 &
  17.82972 \\  
 &
  MPS 1-Layered VQC &
  \multicolumn{1}{c}{20.00\%$\pm$0.10\%} &
  \multicolumn{1}{c}{50.00\%$\pm$1.51\%} &
  \multicolumn{1}{c}{29.00\%$\pm$2.79\%} &
  \multicolumn{1}{c}{41.00\%$\pm$1.98\%} &
  \multicolumn{1}{c}{50.00\%$\pm$2.49\%} &
  \multicolumn{1}{c}{0.00\%$\pm$0.00\%} &
  0.00\%$\pm$0.00\% &
  2733.8904 &
  11.88648 \\  
 &
  MPS 2-Layered VQC &
  \multicolumn{1}{c}{20.00\%$\pm$0.17\%} &
  \multicolumn{1}{c}{50.00\%$\pm$3.65\%} &
  \multicolumn{1}{c}{29.00\%$\pm$1.44\%} &
  \multicolumn{1}{c}{41.00\%$\pm$2.70\%} &
  \multicolumn{1}{c}{50.00\%$\pm$2.00\%} &
  \multicolumn{1}{c}{0.00\%$\pm$7.21\%} &
  0.00\%$\pm$0.00\% &
  2793.3228 &
  17.82972 \\  
 &
  MPS 3-Layered VQC &
  \multicolumn{1}{c}{20.00\%$\pm$0.11\%} &
  \multicolumn{1}{c}{50.00\%$\pm$0.79\%} &
  \multicolumn{1}{c}{29.00\%$\pm$0.40\%} &
  \multicolumn{1}{c}{41.00\%$\pm$0.95\%} &
  \multicolumn{1}{c}{50.00\%$\pm$0.67\%} &
  \multicolumn{1}{c}{0.00\%$\pm$0.00\%} &
  0.00\%$\pm$0.00\% &
  3328.2144 &
  17.82972 \\  
 &
  MPS 4-Layered VQC &
  \multicolumn{1}{c}{80.00\%$\pm$0.19\%} &
  \multicolumn{1}{c}{50.00\%$\pm$0.92\%} &
  \multicolumn{1}{c}{38.00\%$\pm$0.52\%} &
  \multicolumn{1}{c}{60.00\%$\pm$0.29\%} &
  \multicolumn{1}{c}{50.45\%$\pm$0.96\%} &
  \multicolumn{1}{c}{7.30\%$\pm$0.97\%} &
  1.06\%$\pm$0.81\% &
  3209.3496 &
  11.88648 \\  
 &
  TTN 1-Layered VQC &
  \multicolumn{1}{c}{30.00\%$\pm$0.82\%} &
  \multicolumn{1}{c}{50.00\%$\pm$0.50\%} &
  \multicolumn{1}{c}{37.00\%$\pm$0.36\%} &
  \multicolumn{1}{c}{59.00\%$\pm$0.46\%} &
  \multicolumn{1}{c}{50.00\%$\pm$0.47\%} &
  \multicolumn{1}{c}{0.00\%$\pm$0.00\%} &
  0.00\%$\pm$0.00\% &
  3328.2144 &
  11.88648 \\  
 &
  TTN 2-Layered VQC &
  \multicolumn{1}{c}{20.00\%$\pm$0.03\%} &
  \multicolumn{1}{c}{50.00\%$\pm$0.31\%} &
  \multicolumn{1}{c}{29.00\%$\pm$0.85\%} &
  \multicolumn{1}{c}{41.00\%$\pm$0.69\%} &
  \multicolumn{1}{c}{50.00\%$\pm$0.06\%} &
  \multicolumn{1}{c}{0.00\%$\pm$0.00\%} &
  0.00\%$\pm$0.00\% &
  3031.0524 &
  11.88648 \\  
 &
  TTN 3-Layered VQC &
  \multicolumn{1}{c}{30.00\%$\pm$0.59\%} &
  \multicolumn{1}{c}{50.00\%$\pm$0.52\%} &
  \multicolumn{1}{c}{37.00\%$\pm$0.93\%} &
  \multicolumn{1}{c}{59.00\%$\pm$0.98\%} &
  \multicolumn{1}{c}{50.00\%$\pm$0.52\%} &
  \multicolumn{1}{c}{0.00\%$\pm$0.00\%} &
  0.00\%$\pm$0.00\% &
  2436.7284 &
  17.82972 \\  
 &
  TTN 4-Layered VQC &
  \multicolumn{1}{c}{30.00\%$\pm$0.53\%} &
  \multicolumn{1}{c}{50.00\%$\pm$0.39\%} &
  \multicolumn{1}{c}{37.00\%$\pm$0.26\%} &
  \multicolumn{1}{c}{59.00\%$\pm$1.00\%} &
  \multicolumn{1}{c}{50.00\%$\pm$0.05\%} &
  \multicolumn{1}{c}{0.00\%$\pm$0.00\%} &
  0.00\%$\pm$0.00\% &
  2912.1876 &
  17.82972 \\ \hline
\multirow{6}{*}{Hybrid Models} &
  BiLSTMKANnet 1-Layered &
  \multicolumn{1}{c}{88.00\%$\pm$0.46\%} &
  \multicolumn{1}{c}{87.00\%$\pm$0.99\%} &
  \multicolumn{1}{c}{88.00\%$\pm$0.28\%} &
  \multicolumn{1}{c}{88.00\%$\pm$0.90\%} &
  \multicolumn{1}{c}{95.03\%$\pm$0.61\%} &
  \multicolumn{1}{c}{75.12\%$\pm$0.79\%} &
  75.10\%$\pm$0.47\% &
  152.3545 &
  1.2454 \\  
 &
  QCKANnet 1-Layered &
  \multicolumn{1}{c}{83.00\%$\pm$0.87\%} &
  \multicolumn{1}{c}{83.00\%$\pm$0.38\%} &
  \multicolumn{1}{c}{83.00\%$\pm$0.59\%} &
  \multicolumn{1}{c}{83.00\%$\pm$0.88\%} &
  \multicolumn{1}{c}{90.57\%$\pm$0.55\%} &
  \multicolumn{1}{c}{66.10\%$\pm$0.18\%} &
  65.92\%$\pm$0.04\% &
  992.4243 &
  6.2844 \\  
 &
  QCKANnet 2-Layered &
  \multicolumn{1}{c}{82.00\%$\pm$0.09\%} &
  \multicolumn{1}{c}{83.00\%$\pm$0.16\%} &
  \multicolumn{1}{c}{83.00\%$\pm$0.99\%} &
  \multicolumn{1}{c}{83.00\%$\pm$0.97\%} &
  \multicolumn{1}{c}{91.01\%$\pm$0.13\%} &
  \multicolumn{1}{c}{65.82\%$\pm$0.77\%} &
  65.42\%$\pm$0.12\% &
  987.2375 &
  7.3565 \\  
 &
  QCKANnet 3-Layered &
  \multicolumn{1}{c}{81.00\%$\pm$0.53\%} &
  \multicolumn{1}{c}{82.00\%$\pm$0.39\%} &
  \multicolumn{1}{c}{81.00\%$\pm$0.48\%} &
  \multicolumn{1}{c}{82.00\%$\pm$0.68\%} &
  \multicolumn{1}{c}{90.70\%$\pm$0.18\%} &
  \multicolumn{1}{c}{63.13\%$\pm$0.41\%} &
  62.96\%$\pm$0.31\% &
  988.6254 &
  9.4456 \\  
 &
  QCKANnet 4-Layered &
  \multicolumn{1}{c}{83.00\%$\pm$0.18\%} &
  \multicolumn{1}{c}{83.00\%$\pm$0.15\%} &
  \multicolumn{1}{c}{83.00\%$\pm$0.62\%} &
  \multicolumn{1}{c}{83.00\%$\pm$0.04\%} &
  \multicolumn{1}{c}{90.76\%$\pm$0.99\%} &
  \multicolumn{1}{c}{66.10\%$\pm$0.97\%} &
  65.92\%$\pm$0.65\% &
  989.4765 &
  11.5145 \\  
 &
  QDenseKANnet 1-Layered &
  \multicolumn{1}{c}{87.00\%$\pm$0.17\%} &
  \multicolumn{1}{c}{88.00\%$\pm$0.90\%} &
  \multicolumn{1}{c}{88.00\%$\pm$0.17\%} &
  \multicolumn{1}{c}{88.00\%$\pm$0.63\%} &
  \multicolumn{1}{c}{93.53\%$\pm$0.44\%} &
  \multicolumn{1}{c}{75.43\%$\pm$0.74\%} &
  75.38\%$\pm$0.36\% &
  1883.9142 &
  8.3985 \\ \hline
\multirow{4}{*}{Proposed} &
  \textbf{KACQ-DCNN 1-Layered} &
  \multicolumn{1}{c}{\textbf{92.00\%$\pm$0.19}\%} &
  \multicolumn{1}{c}{\textbf{92.00\%$\pm$0.10}\%} &
  \multicolumn{1}{c}{\textbf{92.00\%$\pm$0.07}\%} &
  \multicolumn{1}{c}{\textbf{92.03\%$\pm$0.19}\%} &
  \multicolumn{1}{c}{\textbf{94.77\%$\pm$0.43}\%} &
  \multicolumn{1}{c}{\textbf{83.85\%$\pm$0.81}\%} &
  \textbf{83.84\%$\pm$0.43\%} &
  497.6434 &
  10.489 \\  
 &
  KACQ-DCNN 2-Layered &
  \multicolumn{1}{c}{90.00\%$\pm$0.46\%} &
  \multicolumn{1}{c}{90.00\%$\pm$0.19\%} &
  \multicolumn{1}{c}{90.00\%$\pm$0.32\%} &
  \multicolumn{1}{c}{90.00\%$\pm$0.11\%} &
  \multicolumn{1}{c}{94.00\%$\pm$0.08\%} &
  \multicolumn{1}{c}{80.22\%$\pm$0.36\%} &
  80.22\%$\pm$0.20\% &
  435.4234 &
  11.5411 \\  
 &
  KACQ-DCNN 3-Layered &
  \multicolumn{1}{c}{91.00\%$\pm$0.02\%} &
  \multicolumn{1}{c}{90.00\%$\pm$0.05\%} &
  \multicolumn{1}{c}{90.00\%$\pm$0.03\%} &
  \multicolumn{1}{c}{91.00\%$\pm$0.15\%} &
  \multicolumn{1}{c}{94.00\%$\pm$0.11\%} &
  \multicolumn{1}{c}{80.92\%$\pm$0.30\%} &
  80.84\%$\pm$0.31\% &
  496.6023 &
  13.632 \\  
 &
  KACQ-DCNN 4-Layered &
  \multicolumn{1}{c}{90.00\%$\pm$0.83\%} &
  \multicolumn{1}{c}{90.00\%$\pm$0.38\%} &
  \multicolumn{1}{c}{90.00\%$\pm$0.93\%} &
  \multicolumn{1}{c}{90.00\%$\pm$0.41\%} &
  \multicolumn{1}{c}{94.00\%$\pm$0.17\%} &
  \multicolumn{1}{c}{79.51\%$\pm$0.65\%} &
  79.50\%$\pm$0.14\% &
  639.9443 &
  15.714 \\ \bottomrule[1.5pt]
\end{tabular}%
}
\end{table}

\subsubsection{Performance benchmarking}
We present a comparative evaluation of the performance metrics in Table \ref{tab:perf_compar} across different model categories: CML, quantum neural networks (QNNs), hybrid models, and variants of our proposed KACQ-DCNN.

Among the CML models, CatBoost (86.43\%) outperformed all the other models, showing its effectiveness in handling categorical features and boosting performance. MLP Classifier (85.94\%) closely follows, indicating the potential of neural networks to capture complex patterns. Traditional models such as LR (85.38\%) and KNN (84.33\%) also exhibit competitive performance, emphasizing their simplicity and effectiveness in specific contexts. Tree-based models such as RF (84.27\%) and DT (71.29\%) yield mixed results, with the latter performing poorly because they tend to overfit. Among boosting models, CatBoost and histogram-based gradient boosting (83.80\%) outperform popular alternatives such as XGBoost (83.22\%) and LightGBM (83.77\%), whereas ensemble methods such as AdaBoost (82.11\%) and Bagging (81.55\%) achieve moderate performance. Bernoulli NB (80.58\%) performs well, although the feature independence assumption limits it. Lastly, the dummy classifier (55.44\%) highlights its role as a baseline, far below all other models, indicating the effectiveness of the tested algorithms.

The QNN models demonstrated significantly lower performance across all metrics than their classical counterparts. The MERA 1-Layered VQC achieved the second-highest maP at 53\%, with the MERA 2-Layered VQC showing a sharp decline in performance as the number of layers increased, dropping to 20\%. This substantial reduction underscores the challenges of training deeper quantum models, possibly related to overfitting and optimization difficulties. In contrast, the MPS 4-Layered VQC showed a marginal improvement with the highest 80\% maP and 60\% accuracy, indicating that a balanced architecture is crucial. The inference times for QNN models were notably longer, with some models exceeding 17 seconds, suggesting that while QNNs are still in experimental stages, efficiency remains a critical factor for practical applications.

The hybrid models showcased a blend of classical and quantum techniques. BiLSTMKANnet and QDenseKANnet achieved impressive metrics with an accuracy of 88.00\%, placing them in a favorable position among the CML models and emphasizing the potential of integrating KAN with DL architectures. However, as the number of layers increased in the QCKANnet models, maP exhibited a downward trend up to three layers, highlighting the challenge of maintaining model efficacy with growing complexity. The training and inference times varied considerably, with BiLSTMKANnet requiring $\approx152$ seconds and QDenseKANNet requiring $\approx1884$ seconds for training, highlighting the complexity introduced by the dressed quantum circuits.

The proposed KACQ-DCNN models exhibited exceptional performance across all metrics. KACQ-DCNN 1-Layered achieved the highest accuracy of 92.03\% and an impressive ROC-AUC of 94.77\%, underscoring its robustness and generalizability. Performance metrics remained consistently high across the subsequent layered models, with KACQ-DCNN 3-Layered recording a 91\% accuracy, demonstrating minimal performance degradation with increased complexity. The training times for the KACQ-DCNN models were also reasonable, with the 1-layer model requiring $\approx498$ seconds, showing efficiency in learning compared with the QNN models. 

CML models, particularly CatBoost, MLP classifier, and LR, outperformed quantum models, emphasizing the current limitations of QNNs in practical applications. The drastic performance drop in QNNs as the number of layers increased suggests the need for more sophisticated training techniques and possibly architectural refinements. The hybrid models demonstrated that integrating classical learning approaches with quantum elements can yield promising results, particularly in terms of precision and efficiency. The proposed KACQ-DCNN models set a new benchmark, indicating significant advancements in model design that leverage classical and quantum computing principles to increase performance.

\begin{table}[!ht]
\caption{Comparative analysis of state-of-the-art heart disease prediction models on various datasets}
\label{tab:state-of-the-art}
\resizebox{\columnwidth}{!}{%
\begin{tabular}{lllcccc}
\toprule[1.5pt]
\multicolumn{1}{l}{\multirow{2}{*}{\textbf{Author(s)}}} &
\multicolumn{1}{l}{\multirow{2}{*}{\textbf{Dataset}}} &
\multicolumn{1}{l}{\multirow{2}{*}{\textbf{Model}}} &
\multicolumn{4}{c}{\textbf{Evaluation Metrics}} \\ \cline{4-7} 
\multicolumn{1}{c}{} &
\multicolumn{1}{c}{} &
\multicolumn{1}{c}{} &
\multicolumn{1}{c}{\textbf{maP}} &
\multicolumn{1}{c}{\textbf{maR}} &
\multicolumn{1}{c}{\textbf{maF1}} &
\textbf{Accuracy} \\ \midrule[1pt]
Singh and Kumar~\citep{singh_heart_2020} &
Cleveland dataset &
KNN Classifier (k=3) &
\multicolumn{1}{c}{87.76\%} &
\multicolumn{1}{c}{72.88\%} &
\multicolumn{1}{c}{79.54\%} &
87.00\% \\ 
Kavitha et al.~\citep{kavitha_heart_2021} &
Cleveland dataset &
Hybrid Classifier (Decision Tree + Random Forest &
\multicolumn{1}{c}{-} &
\multicolumn{1}{c}{-} &
\multicolumn{1}{c}{-} &
88.00\% \\ 
Mohan et al.~\citep{mohan_effective_2019} &
Cleveland dataset &
HRFLM (proposed) &
\multicolumn{1}{c}{90.10\%} &
\multicolumn{1}{c}{92.80\%} &
\multicolumn{1}{c}{90.00\%} &
88.40\% \\ 
Sharma et al.~\citep{sharma_heart_2020} &
Cleveland dataset &
Random Forest &
\multicolumn{1}{c}{99.00\%} &
\multicolumn{1}{c}{99.00\%} &
\multicolumn{1}{c}{-} &
- \\ 
Saboor et al.~\citep{saboor_method_2022} &
Cleveland dataset &
SVM Classifier &
\multicolumn{1}{c}{94.00\%} &
\multicolumn{1}{c}{94.00\%} &
\multicolumn{1}{c}{94.00\%} &
96.72\% \\ 
Rani et al.~\citep{rani_decision_2021} &
Cleveland dataset &
Random Forrest &
\multicolumn{1}{c}{88.46\%} &
\multicolumn{1}{c}{84.14\%} &
\multicolumn{1}{c}{86.25\%} &
86.60\% \\ 
Shah et al.~\citep{shah_heart_2020} &
Cleveland dataset &
KNN Classifier (k=7) &
\multicolumn{1}{c}{-} &
\multicolumn{1}{c}{-} &
\multicolumn{1}{c}{-} &
90.78\% \\ 
Li et al.~\citep{li_heart_2020} &
Cleveland dataset &
FCMIM (The feature selector algorithm which led to better results) &
\multicolumn{1}{c}{-} &
\multicolumn{1}{c}{92.50\%} &
\multicolumn{1}{c}{-} &
92.00\% \\ 
Doppala et al.~\citep{doppala_hybrid_2023} &
Cleveland dataset &
Genetic Algorithm + Radial Basis Function &
\multicolumn{1}{c}{95.00\%} &
\multicolumn{1}{c}{96.00\%} &
\multicolumn{1}{c}{95.00\%} &
94.00\% \\ 
Kavitha and Kaulgud~\citep{kavitha_quantum_2023} &
Cleveland dataset &
Quantum K-means &
\multicolumn{1}{c}{96.80\%} &
\multicolumn{1}{c}{95.00\%} &
\multicolumn{1}{c}{96.40\%} &
96.40\% \\ 
Kumar et al.~\citep{kumar_heart_2021} &
Cleveland dataset &
QRFC &
\multicolumn{1}{c}{89.00\%} &
\multicolumn{1}{c}{93.00\%} &
\multicolumn{1}{c}{88.00\%} &
89.00\% \\ 
Babu et al.~\citep{babu_revolutionizing_2024} &
Cleveland dataset &
QuEML &
\multicolumn{1}{c}{93.00\%} &
\multicolumn{1}{c}{94.00\%} &
\multicolumn{1}{c}{93.00\%} &
93.00\% \\ 
Mir et al.~\citep{mir_novel_2024} &
Heart Statlog dataset &
Random Forest + Adaboost &
\multicolumn{1}{c}{99.40\%} &
\multicolumn{1}{c}{99.69\%} &
\multicolumn{1}{c}{99.55\%} &
99.48\% \\ 
Mir et al.~\citep{mir_novel_2024} &
Heart disease UCI dataset &
Random Forest + Adaboost &
\multicolumn{1}{c}{92.23\%} &
\multicolumn{1}{c}{96.75\%} &
\multicolumn{1}{c}{94.44\%} &
93.90\% \\ 
Karadeniz, Tokdemir, and Maraş~\citep{karadeniz_ensemble_2021} &
Heart Statlog dataset &
Shrunk Covariance classifier &
\multicolumn{1}{c}{85.50\%} &
\multicolumn{1}{c}{89.80\%} &
\multicolumn{1}{c}{87.60\%} &
88.80\% \\ 
Anuradha and David~\citep{anuradha_feature_2021} &
Heart Statlog dataset &
Majority Voting Ensemble &
\multicolumn{1}{c}{87.00\%} &
\multicolumn{1}{c}{87.00\%} &
\multicolumn{1}{c}{87.00\%} &
87.04\% \\ 
Patro, Nayak, and Padhy~\citep{patro_heart_2021} &
Heart disease UCI dataset &
Bayesian Optimized Support Vector Machine &
\multicolumn{1}{c}{100.00\%} &
\multicolumn{1}{c}{80.00\%} &
\multicolumn{1}{c}{-} &
93.30\% \\ 
\textbf{Ours} &
\textbf{Cleveland dataset} &
\textbf{Proposed KACQ-DCNN 4-Qubit 1-Layered} &
\multicolumn{1}{c}{\textbf{97.00\%}} &
\multicolumn{1}{c}{\textbf{96.00\%}} &
\multicolumn{1}{c}{\textbf{96.00\%}} &
\textbf{96.23\%} \\ 
\textbf{Ours} &
\textbf{Heart Statlog dataset} &
\textbf{Proposed KACQ-DCNN 4-Qubit 1-Layered} &
\multicolumn{1}{c}{\textbf{97.00\%}} &
\multicolumn{1}{c}{\textbf{96.00\%}} &
\multicolumn{1}{c}{\textbf{96.00\%}} &
\textbf{96.30\%} \\ 
\textbf{Ours} &
\textbf{Heart disease UCI dataset + Heart Statlog dataset} &
\textbf{Proposed KACQ-DCNN 4-Qubit 1-Layered} &
\multicolumn{1}{c}{\textbf{92.00\%}} &
\multicolumn{1}{c}{\textbf{92.00\%}} &
\multicolumn{1}{c}{\textbf{92.00\%}} &
\textbf{92.03\%} \\ \bottomrule[1.5pt]
\end{tabular}%
}
\end{table}

In contrast to the existing literature, mentioned in Table {\ref{tab:state-of-the-art}}, where models were trained on individual heart disease datasets, our approach is novel in that we merged multiple datasets, including the Statlog and UCI datasets, to show the generalizability and robustness of KACQ-DCNN. This is the first attempt in the literature to train a model using such a merged heart disease dataset, offering a more holistic view across different datasets. It's important to highlight that the Cleveland dataset is a small subset of the UCI dataset (Cleveland, Hungarian, Switzerland, and Long Beach, Virginia), on which several studies trained models individually. Our KAN-enhanced method allows for broader model generalization, potentially capturing patterns across datasets, which could improve robustness in real-world applications. Despite this complexity, our proposed KACQ-DCNN achieved 92\% across all evaluation metrics, indicating strong performance even when dealing with the diversity of combined datasets. In the literature, dataset-specific models were used to achieve high accuracy using only one dataset. For example, {\cite{mir_novel_2024}} got 99.48\% accuracy via RF + Adaboost on the Heart Statlog dataset. However, KACQ-DCNN also achieved superior state-of-the-art performance on the Cleveland and Statlog datasets, both sharing an equal maF1 of 96\% and the accuracy of 96.23\% and 96.30\%, respectively.

\begin{figure*}[!ht]
\centering
\begin{subfigure}{0.45\linewidth}\centering
    \includegraphics[width=\linewidth]{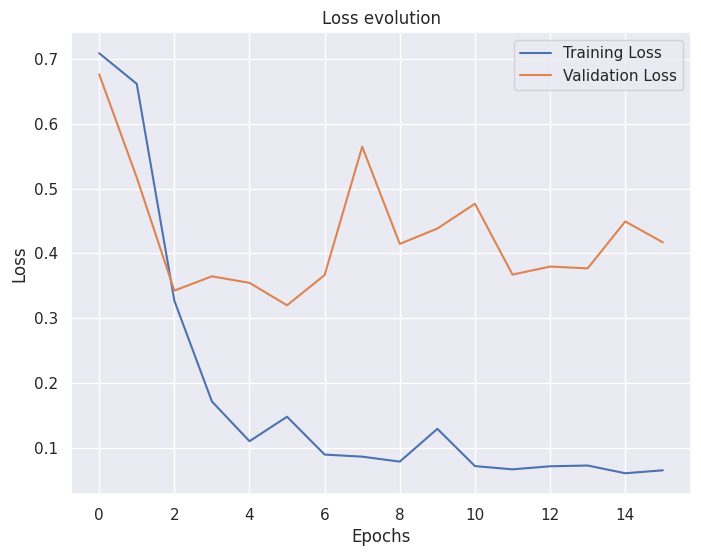}
    \caption{Loss versus epoch plot}
    \label{fig:loss}
\end{subfigure}
\hfill
\begin{subfigure}{0.45\linewidth}\centering
    \includegraphics[width=\linewidth]{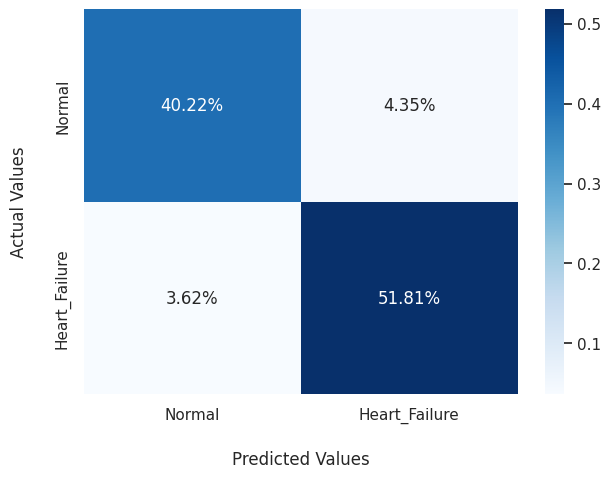}
    \caption{Confusion matrix}
    \label{fig:cm}
\end{subfigure}
\caption{(a) Loss versus epoch curves for both training and validation, and (b) confusion matrix of KACQ-DCNN.}
\label{fig:kacq_performance1}
\end{figure*}

Over 16 epochs, KACQ-DCNN initially showed consistent improvements in terms of loss and accuracy. The first epoch resulted in a loss of 0.7087, with a validation accuracy of 73.55\%. By the third epoch, validation accuracy reached 88.77\%, and by the seventh epoch, it peaked at 92.03\%. The accuracy subsequently fluctuated without significant improvements. The learning rate was reduced in epoch 12 to enhance convergence from $1\times10^{-3}$ to $5\times10^{-4}$. However, validation accuracy could not surpass the 92.03\% peak, and training was halted at epoch 16 owing to the early stopping mechanism. As shown in Figure \ref{fig:loss}, the training loss steadily decreased across all epochs, indicating effective learning. However, the validation loss plateaued around epoch 7, with minor fluctuations, suggesting that the model was near its generalization capacity. Despite the learning rate reduction at epoch 12, validation loss showed no substantial gains, confirming the onset of overfitting. Figure {\ref{fig:cm}} presents the confusion matrix for the KACQ-DCNN model on the test set, offering a granular view of its classification performance. The matrix reveals that 40.22\% of all test instances were correctly identified as 'Normal' (True Negatives), and 51.81\% were correctly identified as `Heart Failure' (TPs). These values indicate strong predictive capability for both classes. The model misclassified 4.35\% of `Normal' instances as `Heart Failure' (FPs), and 3.62\% of `Heart Failure' instances as `Normal' (FNs). The relatively low FN rate is particularly important in a clinical setting, as it signifies fewer missed diagnoses of heart failure. The FP rate, while also low, indicates instances where healthy individuals might be flagged for further investigation.

\begin{figure*}[!ht]
\centering
\begin{subfigure}{0.48\linewidth} \centering
    \includegraphics[width=\linewidth]{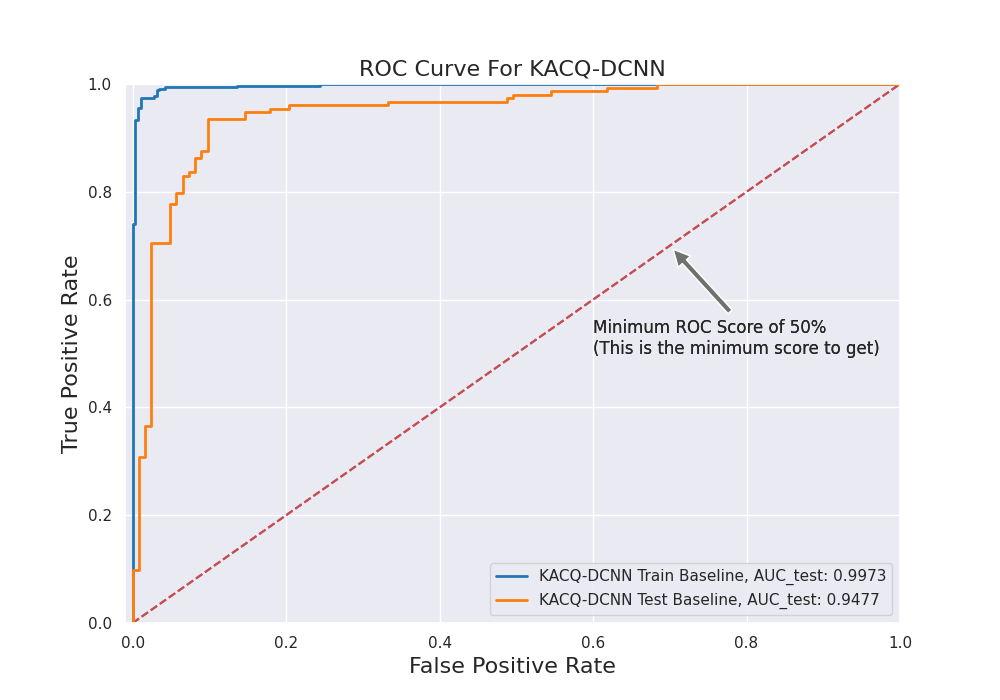}
    \caption{ROC plot for KACQ-DCNN}
    \label{fig:roc_curve}
\end{subfigure}
\hfill
\begin{subfigure}{0.48\linewidth} \centering
    \includegraphics[width=\linewidth]{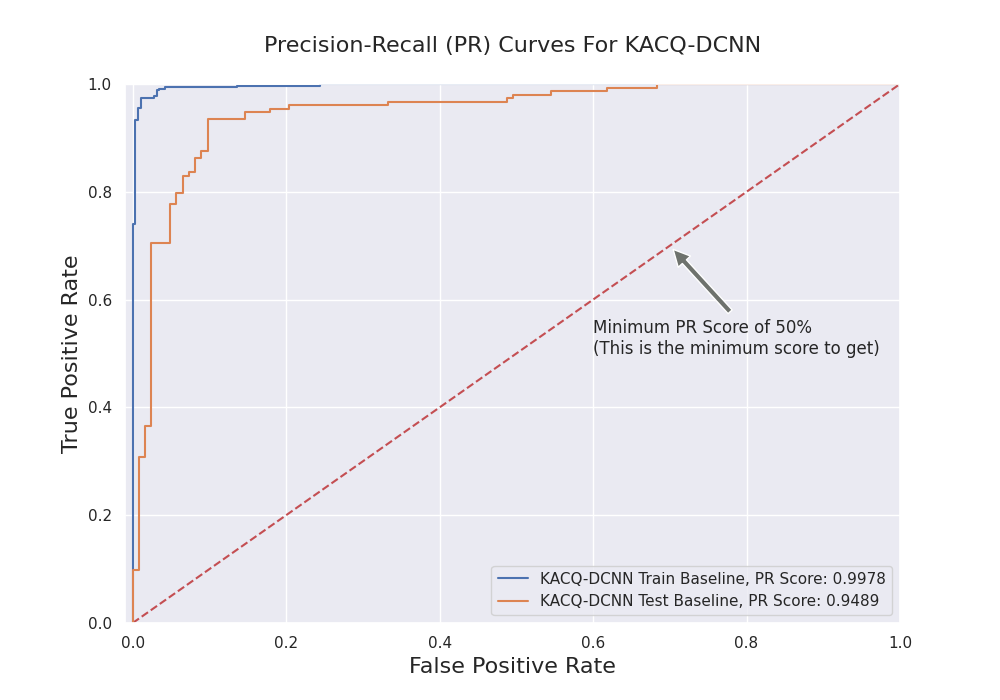}
    \caption{Precision-Recall plot for KACQ-DCNN}
    \label{fig:pr_curve}
\end{subfigure}
\caption{ROC curve and precision-recall curve for the performance of KACQ-DCNN on the training and test datasets.}
\label{fig:kacq_performance2}
\end{figure*}

Figures \ref{fig:roc_curve} and \ref{fig:pr_curve} illustrate the performance of KACQ-DCNN via the ROC and precision-recall curves, which are critical metrics for evaluating binary classification tasks. The ROC curve demonstrates a high TP rate against the FP rate, indicating that the model effectively distinguishes between the positive and negative classes. An AUC score of 99.73\% on the training set and 94.77\% on the test set suggested excellent classification ability, with minimal risk of overfitting. Similarly, the precision-recall curve reveals the model's ability to maintain high precision and recall rates, achieving values of 99.78\% in training and 94.89\% in testing. This finding indicates that the model is accurate and robust across both datasets, demonstrating its effectiveness in predicting the target classes.

\begin{table}[!ht]
\caption{Ablation studies of the proposed KACQ-DCNN. Here, ``w/o'' means without certain components to analyze their impact on performance.}
\label{tab:ablation}
\resizebox{\columnwidth}{!}{%
\begin{tabular}{l ccccccc c c}
\toprule[1.5pt]
\multicolumn{1}{c}{\multirow{2}{*}{\textbf{Models}}} &
  \multicolumn{7}{c}{\textbf{Evaluation Metrics}} &
  \multirow{2}{*}{\textbf{Training Time (s)}} &
  \multirow{2}{*}{\textbf{Inference Time (s)}} \\ \cline{2-8}
\multicolumn{1}{c}{} &
  \multicolumn{1}{c}{\textbf{maP}} &
  \multicolumn{1}{c}{\textbf{maR}} &
  \multicolumn{1}{c}{\textbf{maF1}} &
  \multicolumn{1}{c}{\textbf{Accuracy}} &
  \multicolumn{1}{c}{\textbf{ROC-AUC}} &
  \multicolumn{1}{c}{\textbf{MCC}} &
  \textbf{Kappa} &
   &
   \\ \midrule[1pt]
MLP version of KACQ-DCNN &
  \multicolumn{1}{c}{90.00\%$\pm$0.62\%} &
  \multicolumn{1}{c}{89.00\%$\pm$0.21\%} &
  \multicolumn{1}{c}{90.00\%$\pm$0.70\%} &
  \multicolumn{1}{c}{90.00\%$\pm$0.75\%} &
  \multicolumn{1}{c}{\textbf{95.11\%$\pm$0.68\%}} &
  \multicolumn{1}{c}{79.47\%$\pm$0.88\%} &
  79.34\%$\pm$0.66\% &
  409.20 &
  8.37 \\ 
Replaced BiLSTMs with LSTMs &
  \multicolumn{1}{c}{89.00\%$\pm$0.07\%} &
  \multicolumn{1}{c}{89.00\%$\pm$0.96\%} &
  \multicolumn{1}{c}{89.00\%$\pm$0.36\%} &
  \multicolumn{1}{c}{89.00\%$\pm$0.96\%} &
  \multicolumn{1}{c}{94.20\%$\pm$0.01\%} &
  \multicolumn{1}{c}{77.96\%$\pm$0.88\%} &
  77.93\%$\pm$0.66\% &
  161.37 &
  5.24 \\ 
w/o one BiLSTM layer &
  \multicolumn{1}{c}{89.00\%$\pm$0.81\%} &
  \multicolumn{1}{c}{89.00\%$\pm$0.98\%} &
  \multicolumn{1}{c}{89.00\%$\pm$0.82\%} &
  \multicolumn{1}{c}{89.00\%$\pm$0.48\%} &
  \multicolumn{1}{c}{\textbf{95.09\%$\pm$0.23\%}} &
  \multicolumn{1}{c}{78.72\%$\pm$0.93\%} &
  78.72\%$\pm$0.95\% &
  184.51 &
  5.25 \\ 
w/o all BiLSTM layers &
  \multicolumn{1}{c}{89.00\%$\pm$0.75\%} &
  \multicolumn{1}{c}{89.00\%$\pm$0.19\%} &
  \multicolumn{1}{c}{89.00\%$\pm$0.14\%} &
  \multicolumn{1}{c}{89.00\%$\pm$0.79\%} &
  \multicolumn{1}{c}{\textbf{95.56\%$\pm$0.68\%}} &
  \multicolumn{1}{c}{78.80\%$\pm$0.61\%} &
  78.78\%$\pm$0.02\% &
  251.80 &
  6.29 \\ 
w/o DenseKAN layers in classical channel &
  \multicolumn{1}{c}{89.00\%$\pm$0.94\%} &
  \multicolumn{1}{c}{88.00\%$\pm$0.15\%} &
  \multicolumn{1}{c}{88.00\%$\pm$0.68\%} &
  \multicolumn{1}{c}{88.00\%$\pm$0.97\%} &
  \multicolumn{1}{c}{93.65\%$\pm$0.78\%} &
  \multicolumn{1}{c}{76.68\%$\pm$0.68\%} &
  76.27\%$\pm$0.66\% &
  289.20 &
  9.45 \\ 
w/o DenseKAN layers in quantum channel &
  \multicolumn{1}{c}{91.00\%$\pm$0.53\%} &
  \multicolumn{1}{c}{91.00\%$\pm$0.31\%} &
  \multicolumn{1}{c}{91.00\%$\pm$0.18\%} &
  \multicolumn{1}{c}{91.00\%$\pm$0.25\%} &
  \multicolumn{1}{c}{93.84\%$\pm$0.15\%} &
  \multicolumn{1}{c}{81.68\%$\pm$0.95\%} &
  81.68\%$\pm$0.84\% &
  385.60 &
  8.39 \\ 
w/o quantum layers &
  \multicolumn{1}{c}{89.00\%$\pm$0.31\%} &
  \multicolumn{1}{c}{89.00\%$\pm$0.15\%} &
  \multicolumn{1}{c}{89.00\%$\pm$0.07\%} &
  \multicolumn{1}{c}{89.00\%$\pm$0.21\%} &
  \multicolumn{1}{c}{\textbf{95.52\%$\pm$0.95\%}} &
  \multicolumn{1}{c}{78.04\%$\pm$0.91\%} &
  77.82\%$\pm$0.78\% &
  40.88 &
  0.16 \\ 
w/o dropout layers &
  \multicolumn{1}{c}{90.00\%$\pm$0.76\%} &
  \multicolumn{1}{c}{90.00\%$\pm$0.79\%} &
  \multicolumn{1}{c}{90.00\%$\pm$0.54\%} &
  \multicolumn{1}{c}{90.00\%$\pm$0.27\%} &
  \multicolumn{1}{c}{94.51\%$\pm$0.78\%} &
  \multicolumn{1}{c}{80.19\%$\pm$0.42\%} &
  80.18\%$\pm$0.69\% &
  396.43 &
  7.34 \\ 
KACQ-DCNN with Angle Embedding &
  \multicolumn{1}{c}{88.00\%$\pm$0.88\%} &
  \multicolumn{1}{c}{88.00\%$\pm$0.07\%} &
  \multicolumn{1}{c}{88.00\%$\pm$0.65\%} &
  \multicolumn{1}{c}{88.00\%$\pm$0.46\%} &
  \multicolumn{1}{c}{91.48\%$\pm$0.55\%} &
  \multicolumn{1}{c}{75.95\%$\pm$0.82\%} &
  75.90\%$\pm$0.31\% &
  406.50 &
  10.47 \\ 
\textbf{KAN version of KACQ-DCNN (1-Layered)} &
  \multicolumn{1}{c}{\textbf{92.00\%$\pm$0.19\%}} &
  \multicolumn{1}{c}{\textbf{92.00\%$\pm$0.10\%}} &
  \multicolumn{1}{c}{\textbf{92.00\%$\pm$0.07\%}} &
  \multicolumn{1}{c}{\textbf{92.03\%$\pm$0.19}\%} &
  \multicolumn{1}{c}{94.77\%$\pm$0.43\%} &
  \multicolumn{1}{c}{\textbf{83.85\%$\pm$0.81}\%} &
  \textbf{83.84\%$\pm$0.43\%} &
  497.64 &
  10.49 \\ \bottomrule[1.5pt]
\end{tabular}%
}
\end{table}

Key observations from Table{~\ref{tab:perf_compar}} regarding overall performance benchmarking include:
\begin{enumerate}
    \item The proposed KACQ-DCNN 1-Layered model (Accuracy: 92.03\%, ROC-AUC: 94.77\%) significantly outperforms all 37 benchmark models, including CML, QNN, and other hybrid variants.
    \item Among classical models, CatBoost (86.43\% accuracy) demonstrates the best performance, highlighting its effectiveness with categorical features. MLP classifiers also show strong results.
    \item Pure QNN models generally exhibit substantially lower performance and longer inference times compared to classical and hybrid KACQ-DCNN models, indicating current challenges in their practical application for this task. Performance often degraded with increased quantum layers.
    \item Hybrid models like BiLSTMKANnet and QDenseKANnet (both achieving 88.00\% accuracy) show promise, outperforming many CML models and bridging the gap towards the more advanced KACQ-DCNN.
\end{enumerate}

\subsubsection{Ablation study analysis}
The ablation study of the proposed KACQ-DCNN model, as shown in Table \ref{tab:ablation}, involves the selective modification, removal, or isolation of various components to analyze their contributions to overall performance. 

While the MLP version of KACQ-DCNN serves as the baseline, achieving competitive scores across all metrics with an accuracy of 90\%, it still lags behind the KAN version of KACQ-DCNN. This shows that with the same number of parameters, KAN-based KACQ-DCNN outperforms MLP-based KACQ-DCNN, which denotes the ability of our model to perform well in resource-constrained environments. However, it took significantly less time to train than the KAN-based KACQ-DCNN. Notably, replacing BiLSTMs with LSTMs resulted in a drop in accuracy to 89\%, suggesting that the bidirectional processing capability of the BiLSTM layers is beneficial for capturing context in sequential data. Further analysis revealed that removing one BiLSTM layer retained the accuracy of 89\%, indicating that the model's robustness was partially maintained even with reduced complexity. However, omitting all BiLSTM layers yielded an accuracy of 89\%, suggesting a possible redundancy in specific configurations; however, its striking ROC-AUC improvement from 94.20\% to 95.56\% underscores the importance of two BiLSTM layers in KACQ-DCNN. The DenseKAN layers in the classical channel were crucial because their removal led to a decline in maF1 and accuracy to 88\%. Conversely, omitting DenseKAN layers in the quantum channel surprisingly resulted in improved accuracy (91\%), suggesting that these layers may introduce complexity that the model can navigate without.

The absence of quantum layers had a dramatic effect, with the accuracy dropping back to 89\%; MCC and kappa also significantly decreased, although the ROC-AUC increased to 95.52\%. This finding indicates that quantum components significantly contribute to performance. Furthermore, removing the dropout layers did not adversely affect the accuracy, maintaining a 90\% score, which emphasizes the model's stability under various configurations; it helped MCC and kappa increase to the top third. It also required less training and inference time than KACQ-DCNN because of lower computational overhead, although the better performance of KACQ-DCNN necessitates the presence of dropout layers. Interestingly, replacing AE with `angle embedding' to the KACQ-DCNN framework hurt performance, with accuracy dropping to 88\%. This suggests that the change may make the model more difficult to understand or use for decision-making.

The KACQ-DCNN 1-layered configuration achieved the highest accuracy of 92.03\%, coupled with superior scores across other metrics, except for a negligible performance drop in ROC-AUC. It also took most of the training time, but the inference time differed slightly for all models. This underscores the potential for simplified architectures to perform exceptionally well while highlighting the need to balance complexity and performance. Overall, this ablation study effectively illustrates the nuanced role of each component within the KACQ-DCNN architecture, providing valuable insights for future enhancements and optimizations.

The key observations from the ablation study (Table{~\ref{tab:ablation}}) are:
\begin{enumerate}
\item The KAN-version of KACQ-DCNN (92.03\% accuracy) substantially outperforms the MLP version (90.00\% accuracy) with a similar parameter count, underscoring KAN's superior function approximation capabilities.
\item Replacing BiLSTMs with LSTMs or removing BiLSTM layers led to a noticeable drop in accuracy (to 89\%), indicating the importance of bidirectional sequential processing. However, the model maintained some robustness.
\item Removing DenseKAN layers from the classical channel reduced accuracy to 88\%, highlighting their crucial role.
\item Removing quantum layers reduced accuracy back to 89\% and significantly impacted MCC and Kappa, confirming the positive contribution of the quantum components. Conversely, removing DenseKAN layers from the quantum channel improved accuracy to 91\%, suggesting a simplification in that part of the model was beneficial.
\item Using angle embedding instead of amplitude embedding for quantum circuits decreased accuracy to 88\%, suggesting AE is more suitable for this architecture.
\item Removing dropout layers maintained 90\% accuracy and reduced training time, but the top-performing KACQ-DCNN includes them, suggesting they help achieve the peak 92.03\% by mitigating overfitting during more extensive optimization.
\end{enumerate}

\subsubsection{Statistical validation}
To statistically validate the performance of our proposed model, KACQ-DCNN 1-Layered, we conducted two-tailed paired t-tests between the accuracy of this model and nine other best-performing benchmarked models obtained by 10-fold cross-validation. The null and alternative hypotheses for the paired t-test are as follows:
\\
\\
\textit{Null Hypothesis} ($H_{0}$): There is no significant difference between the accuracy of the KACQ-DCNN 1-Layered model and the other benchmarked models. Mathematically, this can be stated as:
    \[
    H_{0}: \mu_d = 0
    \]
    where \(\mu_d\) represents the mean difference in accuracy between the KACQ-DCNN 1-Layered model and each benchmark model. 
\\
\\
\textit{Alternative Hypothesis} ($H_{1}$): There is a significant difference in accuracy between the KACQ-DCNN 1-Layered model and the other models. Formally, this can be expressed as follows:
    \[
    H_{1}: \mu_d \neq 0
    \]

\begin{table}[!ht]
\centering
\caption{Stratified 10-fold cross-validated two-tailed paired t-tests between proposed KACQ-DCNN 1-Layered and all the benchmark models}
\label{tab:t-test}
\footnotesize
\begin{tabular}{l c c c c} \toprule[1.5pt]    
\textbf{Model} &  \textbf{Degrees of freedom} & \textbf{Cohen's d} & \textbf{t-statistic (Accuracy)} & \textbf{p-value (Accuracy)} \\ \midrule[1pt] 
CatBoost &  9& 4.979&15.745& 7.40E-08
\\ 
MLP Classifier &  9& 6.235&19.717& 1.03E-08
\\  
Logistic Regression &  9& 4.566&14.438& 1.57E-07
\\  
K-Nearest Neighbor &  9& 6.754&21.357& 5.08E-09
\\  
BiLSTMKANnet 1-Layered &  9& 4.310&13.629& 2.59E-07
\\  
QDenseKANnet 1-Layered &  9& 5.365&16.965& 3.86E-08
\\  
KACQ-DCNN 2-Layered &  9& 9.558&30.225& 2.32E-10
\\  
KACQ-DCNN 3-Layered &  9& 4.802&15.186& 1.01E-07
\\  
KACQ-DCNN 4-Layered &  9& 4.343&13.734& 2.42E-07
\\ \bottomrule[1.5pt]
\end{tabular}
\end{table}

We conducted a two-tailed paired t-test at a significance level of \(\alpha = 0.05\), as shown in Table \ref{tab:t-test}. To account for multiple comparisons, we applied the Bonferroni correction, adjusting the significance level to \(\alpha_{\text{adjusted}} = \frac{\alpha}{9} = 0.0056\) to control the family-wise error rate. The effect size for each comparison is calculated via Cohen's d formula to provide a standardized measure of the magnitude of the difference. Since all p-values are much lower than the adjusted significance level of \(0.0056\), we reject the null hypothesis for all models. This implies a statistically significant difference in accuracy between the KACQ-DCNN 1-Layered model and all the benchmark models. Thus, we can conclude that the proposed KACQ-DCNN 1-Layered model significantly improves accuracy over the other models tested.

Key observations from the statistical validation in Table{~\ref{tab:t-test}} include:
\begin{enumerate}
    \item All calculated p-values for the paired t-tests between KACQ-DCNN 1-Layered and the nine other top-performing models are substantially lower than the Bonferroni-corrected significance level of \(\alpha_{\text{adjusted}} = \frac{\alpha}{9} = 0.0056\).
    \item This allows us to confidently reject $H_{0}$ for all comparisons, indicating a statistically significant difference in accuracy.
    \item Cohen's d values are consistently large (ranging from 4.310 to 9.558), demonstrating that the observed accuracy improvements of KACQ-DCNN 1-Layered are not only statistically significant but also practically meaningful in magnitude.
    \item These results provide robust statistical evidence that the proposed KACQ-DCNN 1-Layered model significantly improves prediction accuracy over the other benchmarked models for heart disease detection on this dataset.
\end{enumerate}

\subsection{Interpretability analysis}\label{sec4.2}
We aimed to improve the transparency and interpretability of our novel KACQ-DCNN model for heart disease detection. We used two well-known XAI techniques: local interpretable model-agnostic explanations (LIME) and Shapley additive explanations (SHAP). These methods offer valuable insights into how our model makes decisions, providing both local and global perspectives on the impact of individual features on predictions. 

\subsubsection{LIME: local interpretability}
LIME is designed to elucidate the behavior of complex models by generating local explanations for individual predictions. It operates by perturbing the input features of a given instance and observing changes in the model's predictions. To achieve this, LIME constructs a simple, interpretable surrogate model, such as linear regression or DT, around perturbed examples. This surrogate model approximates the decision boundary of the complex model in the vicinity of the instance of interest, thus revealing how small changes in feature values affect the model's predictions. The objective function for LIME is as follows:
\begin{equation}
\hat{h} = \underset{h \in \mathcal{H}}{\text{argmin}} \, L(f, h, \pi_x) + \Omega(h)
\end{equation}
where $f$ represents the original complex model, $h$ is the interpretable surrogate model, and $L(f, h, \pi_x)$ is the loss function that quantifies the fidelity of $h$ to $f$ within the neighborhood defined by $\pi_x$. The regularization term $\Omega(h)$ penalizes the complexity of $h$, thereby ensuring that the surrogate model remains simple and interpretable.

\begin{figure}[!ht]
\centering
\includegraphics[width=1\linewidth]{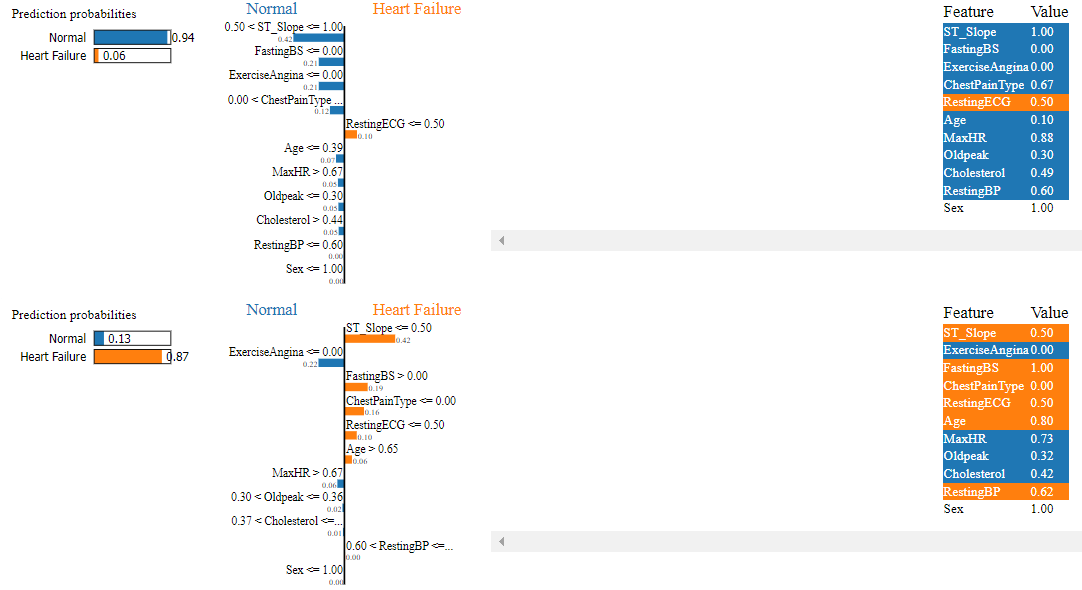}
\caption{Comparison of LIME explanations of KACQ-DCNN for HF predictions: `Normal' vs. `Heart Failure' cases with feature importance visualization and probability breakdown.}
\label{fig:Two_instances}
\end{figure}

Figure \ref{fig:Two_instances} presents two LIME outputs for the HF prediction model, each comprising prediction probabilities, feature importance visualizations, and feature value tables. The top case predicts ``normal'' with 94\% probability, whereas the bottom predicts ``heart failure'' with 87\% probability, allowing examination of how different feature values influence the model's decisions. In both cases, the $ST\_Slope$ appears most influential; for the ``normal'' prediction, a higher $ST\_Slope$ value ($0.50 < \textit{ST\_Slope} \leq 1.00$) strongly indicates a healthy outcome. In contrast, for the ``heart failure'' prediction, a lower $ST\_Slope$ value ($\textit{ST\_Slope} \leq 0.50$) is key. Other important features include $ExerciseAngina$, $FastingBS$, $ChestPainType$, and $RestingECG$, though their impacts vary between cases. The visualization effectively shows how each feature contributes to the prediction, pushing it toward either `Normal' or `Heart Failure.' Feature value tables on the right provide specific values for each case, enabling direct comparison of how these values correspond to the model's predictions.

\begin{figure*}[!ht]
\centering
\begin{subfigure}{0.45\linewidth} \centering
    \includegraphics[width=\linewidth]{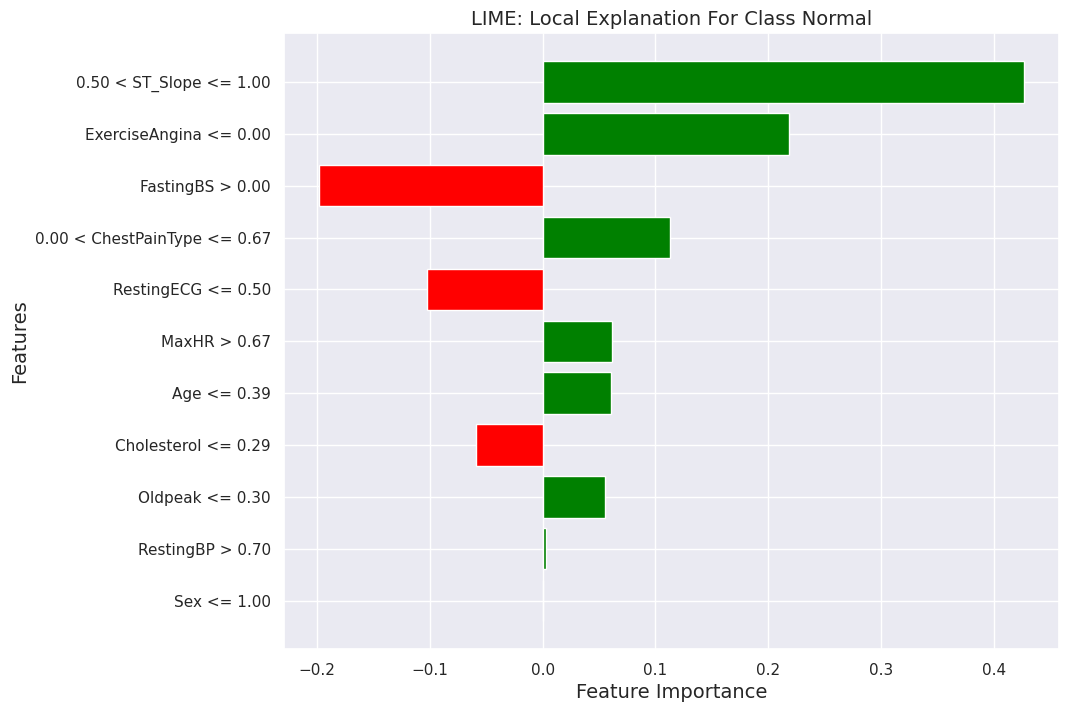}
    \caption{Local explanations of LIME for a normal heart condition instance}
    \label{fig:lime_normal_heart}
\end{subfigure}
\hfill
\begin{subfigure}{0.45\linewidth} \centering
    \includegraphics[width=\linewidth]{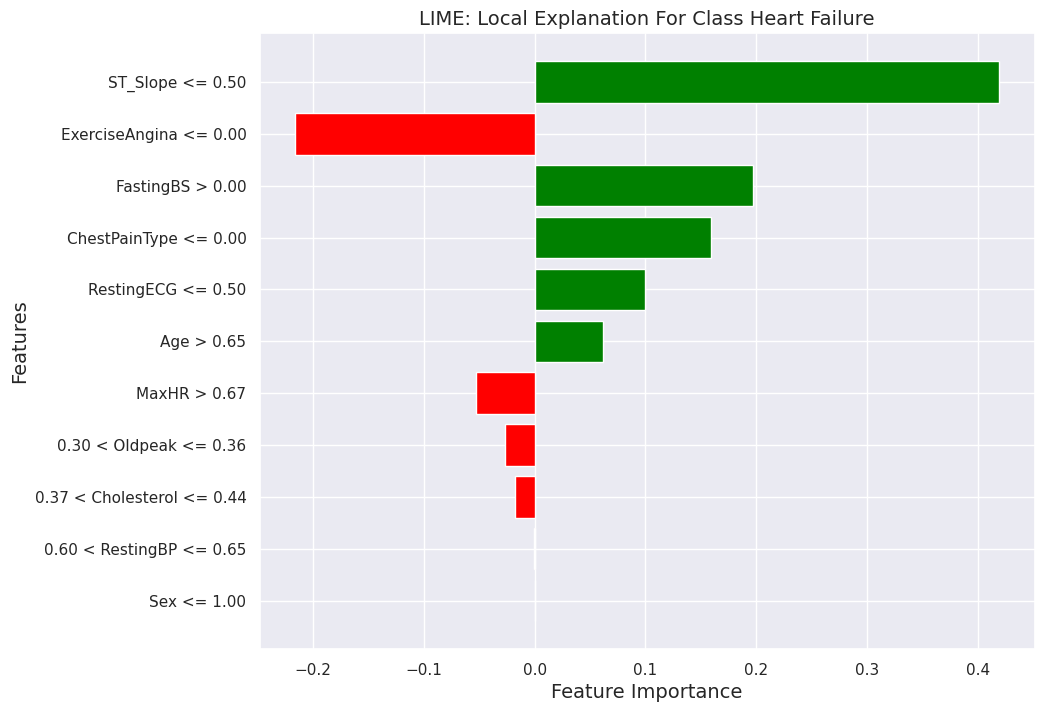}
    \caption{Local explanations of LIME for a heart failure instance.}
    \label{fig:lime_heart_failure}
\end{subfigure}
\caption{LIME local interpretations of KACQ-DCNN for (a) normal heart condition and (b) heart failure cases, illustrating feature contributions and their impact on the model’s predictions.}
\label{fig:lime_combined}
\end{figure*}

Figure \ref{fig:lime_heart_failure} shows the importance of $ST\_Slope$ and $ExerciseAngina$, which are prominent in determining the classification outcome, with contrasting effects between HF and normal classifications. Figure \ref{fig:lime_normal_heart} presents the LIME explanation for predicting a healthy heart condition. Features like $ST\_Slope$ between 0.50 and 1.00, absence of $ExerciseAngina$ ($\leq 0.00$), and values of chest pain type within the range of $0.00 < ChestPainType \leq 0.67$ positively contribute (green bars) to the normal classification. Conversely, certain factors such as $FastingBS$ and $RestingECG$ deviations negatively influence the likelihood of a normal classification (red bars). Figure \ref{fig:lime_heart_failure} shows that the top contributing features towards the classification of HF include a lower ST\_Slope ($\leq 0.50$), absence of $ExerciseAngina$ ($\leq 0.00$), high $FastingBS$ ($> 0.00$), and $ChestPainType$ ($\leq 0.00$). These features have positive contributions (green bars) to HF prediction, with $ST\_Slope$ having the largest impact. Some features, like $ExerciseAngina$ ($\leq 0.00$), have negative contributions (red bars), indicating their absence reduces the likelihood of HF. This comparison between the two LIME visualizations highlights how the same features can have contrasting effects depending on their values and the predicted class.

\subsubsection{SHAP: global interpretability}
SHAP provides a global understanding of model predictions by assigning Shapley values to individual features. These values, derived from cooperative game theory, quantify each feature's contribution to the model's output. SHAP values offer a comprehensive assessment of feature importance and interactions across the entire dataset, enabling us to understand the global behavior of the model. The Shapley value for a feature $j$ can be calculated as:
\begin{equation}
\phi_j = \sum_{S \subseteq N \setminus \{j\}} \frac{|S|!(|N| - |S| - 1)!}{|N|!} \left[ v(S \cup \{j\}) - v(S) \right]
\end{equation}
where $\phi_j$ denotes the Shapley value for feature $j$, $S$ represents a subset of features excluding $j$, $v(S)$ is the model’s prediction for the subset $S$, and $N$ is the total set of features. This formula calculates the average contribution of feature $j$ by summing over all possible subsets of features.

\begin{figure}[!ht]
\centering
\begin{subfigure}{0.45\linewidth} \centering
    \includegraphics[width=\linewidth]{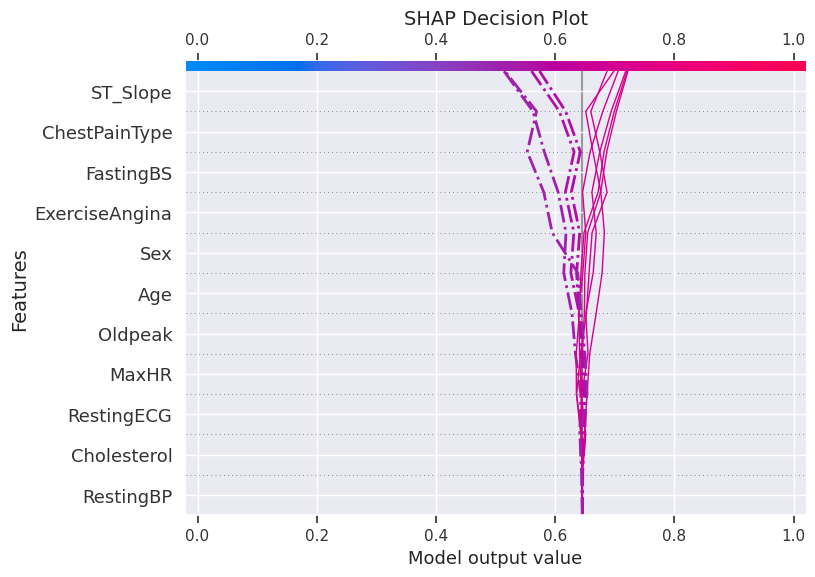}
    \caption{SHAP decision plot}
    \label{fig:shap_decision_plot}
\end{subfigure}
\hfill
\begin{subfigure}{0.45\linewidth} \centering
    \includegraphics[width=\linewidth]{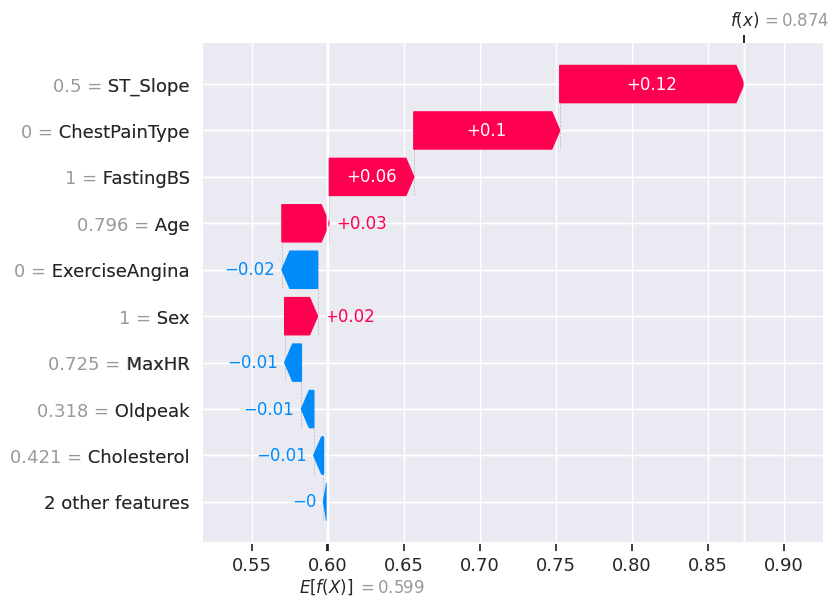}
    \caption{SHAP waterfall plot}
    \label{fig:shap_waterfall_plot}
\end{subfigure}
\caption{(a) SHAP decision plot showing how $ST\_Slope$, $ChestPainType$, and $FastingBS$ influence predictions for KACQ-DCNN, as indicated by the sharp inflection points. The x-axis represents the model's output value, where values closer to 0 indicate a lower probability, and those near 1 suggest a higher probability of the predicted class. The dotted line represents misclassified instances. (b) SHAP waterfall plot illustrating the contribution of individual features to the predicted probability of heart disease. Red bars indicate features that increase the predicted probability, while blue bars represent features that decrease it.}
\label{fig:shap_combined1}
\end{figure}

\begin{figure}[!ht]
\centering
\begin{subfigure}{1\linewidth} \centering
    \includegraphics[width=\linewidth]{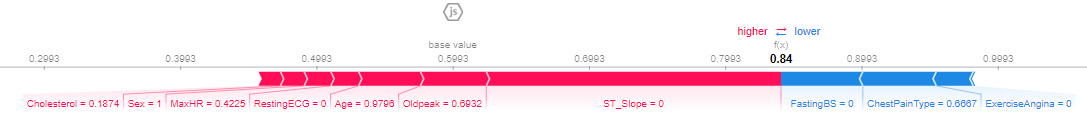}
    \caption{SHAP force plot showing the impact of individual features on KACQ-DCNN's prediction. Red sections represent features that increase the predicted probability, while blue sections represent features that decrease it. The base value is 0.5993, with the final predicted value being 0.84.}
    \label{fig:shap_force_plot}
\end{subfigure}
\hfill
\begin{subfigure}{1\linewidth} \centering
    \includegraphics[width=\linewidth]{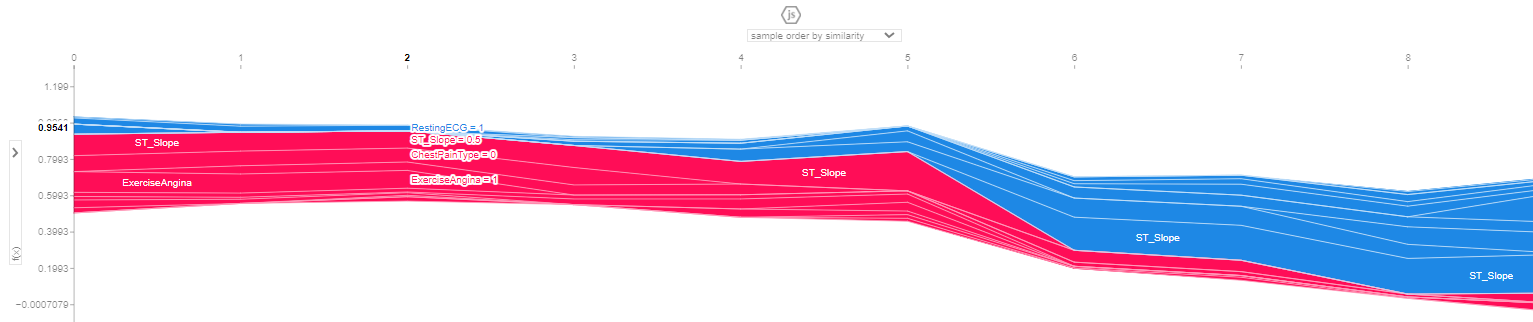}
    \caption{Detailed SHAP summary plot illustrating the cumulative impact of features across multiple samples, ordered by similarity. Each line corresponds to a prediction, and the color indicates the impact of individual features on these predictions. The x-axis represents individual instances, while the y-axis represents the model's predicted values.}
    \label{fig:shap_force_plot2}
\end{subfigure}
\caption{SHAP force plots of KACQ-DCNN for (a) a single instance and (b) multiple instances, demonstrating how different features contribute to the model's output.}
\label{fig:shap_combined2}
\end{figure}

Applying SHAP to our model provided a global perspective on the importance of features and their contributions to predictions. SHAP decision plot (Figure \ref{fig:shap_decision_plot}) further demonstrated how features such as $ST\_Slope$, $ChestPainType$, $FastingBS$, and $ExerciseAngina$ significantly influence the prediction, as indicated by the sharp inflection points. These features push the model output towards a higher prediction score. Other features, such as $RestingECG$ and $FastingBS$ have less impact on the model's final decision, as shown by more minor deviations in the path. SHAP waterfall plot in Figure \ref{fig:shap_waterfall_plot} visualizes the contribution of various features to a prediction made by the model. The base value \( E[f(X)] = 0.599 \),  adjusted by each feature's contribution, represents the average model output, whereas the final model prediction \( f(x) = 0.874 \) is shown at the top. The most influential features include $ST\_Slope$ (+0.12), $ChestPainType$ (+0.1), and $FastingBS$ (+0.06), which all significantly increase the predicted probability. Conversely, features such as $ExerciseAngina$ (-0.02) and $MaxHR$ (-0.01) have negative impacts, slightly reducing the model's prediction. SHAP force plot in Figure \ref{fig:shap_force_plot} visualizes how individual features influence a specific prediction by showing the combined positive and negative contributions of each feature. For this particular prediction, the most impactful positive features are $ST\_Slope$ and $Oldpeak$, which push the prediction upwards towards 0.84. Conversely, features such as $ExerciseAngina$, $ChestPainType$, and $FastingBS$ have a slight negative effect, reducing the prediction. The plot effectively highlights which factors led to the final prediction and in which direction they influenced the outcome. Figure \ref{fig:shap_force_plot2} summarizes multiple predictions, revealing the cumulative effect of features across different samples. For this group of predictions, $ST\_Slope$ and $ExerciseAngina$ emerge as highly influential features. Other features like $ChestPainType$ and $RestingECG$ also play significant roles, but $ST\_Slope$ appears repeatedly as a strong driver in either direction. The red areas show that for many instances, $ST\_Slope$ and $ExerciseAngina$ tend to increase the predicted probability of heart disease, while blue areas for $ST\_Slope$ push it down for others. This global view emphasizes the consistency of these critical features in shaping the model's predictions, reflecting their significant role in cardiac risk assessment.

\subsection{Conformal prediction analysis}\label{sec4.3}
\begin{figure*}[!ht]
\centering
\begin{subfigure}{0.45\linewidth} \centering
    \includegraphics[width=\linewidth]{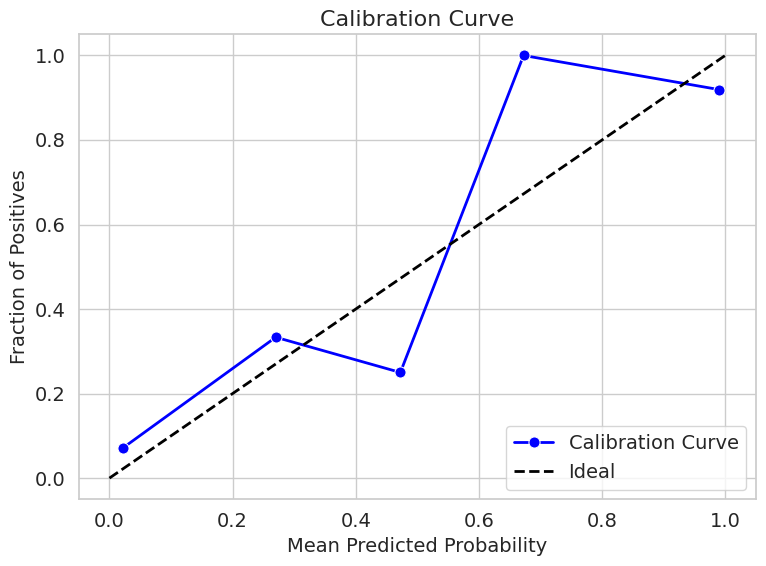}
    \caption{Calibration plot}
    \label{fig:uq1}
\end{subfigure}
\hfill
\begin{subfigure}{0.45\linewidth} \centering
    \includegraphics[width=\linewidth]{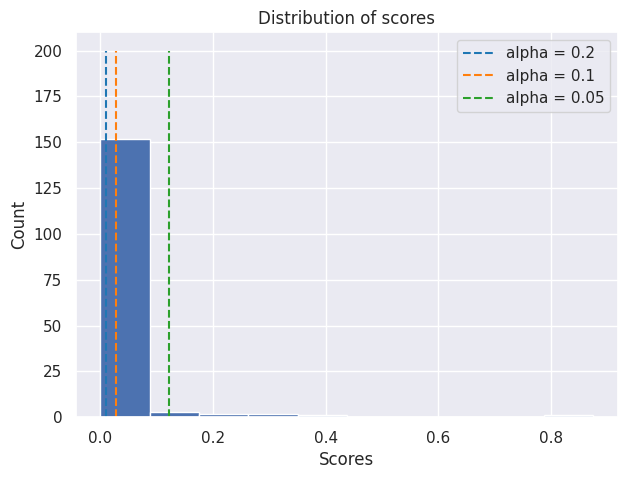}
    \caption{Distribution of the scores}
    \label{fig:uq2}
\end{subfigure}
\hfill
\begin{subfigure}{0.8\linewidth} \centering
    \includegraphics[width=\linewidth]{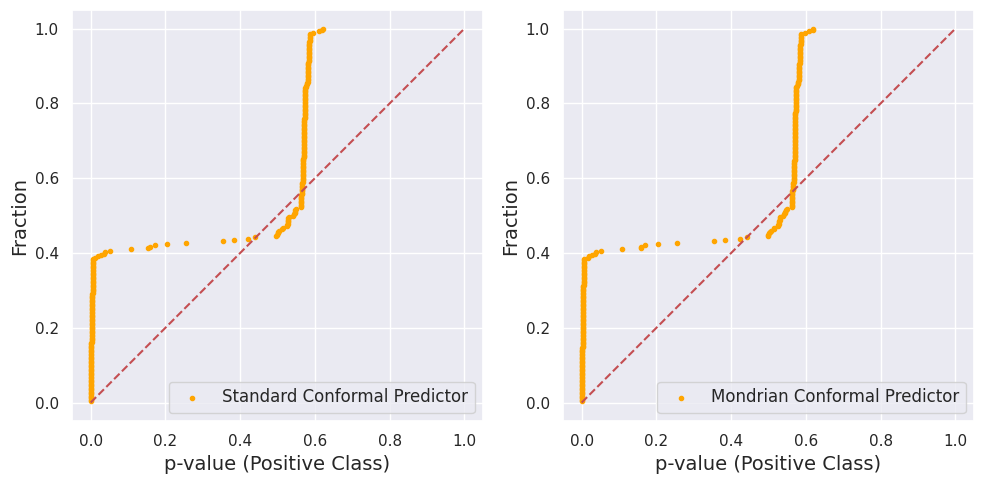}
    \caption{Empirical calibration plots for the Standard conformal predictor and the Mondrian conformal predictor}
    \label{fig:uq3}
\end{subfigure}
\caption{(a) Calibration curve evaluating how well the predicted probabilities from KACQ-DCNN align with the true outcomes. The x-axis represents the mean predicted probability, whereas the y-axis represents the fraction of positives. Ideally, if the model is well-calibrated, the points should lie close to the dashed diagonal line (labeled ``Ideal"), representing perfect calibration. (b) Distribution of conformity scores with quantile thresholds for different confidence levels (\(\alpha = 0.2, 0.1, 0.05\)). The quantile thresholds are overlaid as dashed lines in blue, orange, and green, respectively, showing where the corresponding quantile lies in the distribution. (c) Empirical calibration of the conformal predictors by comparing the sorted p-values for the positive class (``heart failure") against the cumulative fraction of instances. The red dashed line represents the ideal calibration where the predicted p-values perfectly match the observed empirical frequencies, indicating a perfectly calibrated model.}
\label{fig:uq_master1}
\end{figure*}

The calibration curve in Figure \ref{fig:uq1} illustrates the relationship between the predicted probabilities of KACQ-DCNN and the actual fraction of positive outcomes. An ideal calibration curve would follow the diagonal line, indicating that the predicted probabilities align perfectly with the true probabilities. However, as observed, the model exhibits miscalibration in several regions. At lower-mid predicted probabilities (30\%–60\%), the model tends to underestimate the occurrence of positive outcomes, with the curve falling below the ideal line. Conversely, in the upper-mid range of predicted probabilities (60\%–80\%), the model overestimates the likelihood of positive outcomes, as indicated by the curve exceeding the diagonal. This overestimation can result in a greater number of FPs. At lower (0\%–30\%) and higher probabilities (80\%–100\%), the model's predictions are closer to the actual outcomes, although slight overprediction remains. These deviations indicate that the predicted probabilities do not fully represent the true likelihoods, especially in the lower-mid and upper-mid ranges. To address this issue, recalibration methods such as Platt Scaling or Isotonic Regression could be employed to improve the alignment between predicted probabilities and actual outcomes.

Figure \ref{fig:uq3} shows that the Standard conformal predictor (left plot) deviates from the diagonal line, mainly in the middle range of p-values. This suggests that it might not be as confident in its predictions for some ranges of positive class probabilities. Similarly, the Mondrian conformal predictor (right plot) also demonstrates deviations from the ideal line. However, the nature and magnitude of these deviations differ because of the additional conditioning on feature bins (Mondrian approach). The Standard and Mondrian conformal predictors initialized via ``crepes'' were tested on several important metrics, including error rate, average prediction set size, single-class predictions, and number of empty sets. In addition, we recorded the computational efficiency with respect to fitting and evaluation time. These results provide insights into the calibration, precision, and computational feasibility of the model.

The ``error rate'' represents the proportion of instances where the true class was not included in the prediction set. We observed an error rate of 7.5\% for the Standard conformal predictor, whereas for the Mondrian conformal predictor, the error rate was slightly lower at 6.5\%. This means that in only 6.5\% of the test cases, the true class was not part of the predicted set. This suggests that the Mondrian variant offers a marginal improvement in reliability by more consistently including the true label within the prediction set. The ``average size of the prediction set'' indicates the number of labels included for each instance, reflecting the model’s precision. The standard approach yielded an average prediction set size of 1.62, whereas the Mondrian approach reduced this value to 1.53, demonstrating higher confidence in its predictions. A smaller prediction set is generally more desirable, providing more specific and actionable predictions. The proportion of cases where the prediction set contained exactly one label, known as ``single-class predictions,'' was 38\% for the standard method and 47\% for the Mondrian approach. This shows that the Mondrian method is more confident, making definitive predictions for nearly half of the test cases. Neither of the methods produced any ``empty prediction sets'' (0.0), indicating that both models could always produce a valid prediction set. This is a critical outcome, as it ensures that the conformal predictor can always provide an answer, regardless of the difficulty of the classification task. The ``time taken to fit'' the model was 2.408 seconds for the standard conformal predictor and 0.00014 seconds for the Mondrian version. The evaluation times were 0.0048 seconds and 0.0089 seconds, respectively. Both methods exhibited high computational efficiency, with the Mondrian predictor performing slightly faster during fitting and evaluation. This makes them viable for real-time applications where quick predictions are necessary.

We developed a custom wrapper to integrate a TensorFlow-based sequential model with the MAPIE framework for conformal predictions. Early stopping was applied on the basis of validation loss to prevent overfitting. We generated class probability estimates by leveraging the model's outputs, which were then used to form class predictions based on a threshold of 0.5. This TensorFlow-wrapped model was then integrated into MAPIE’s conformal prediction framework to estimate prediction sets with a confidence level \(1 - \alpha\). Specifically, we use sigmoid outputs from the wrapper model as conformity scores, and MAPIE computes the conformal quantile \(\hat{q}\), which helps define the prediction sets for new data points. These prediction sets were designed to include all classes whose sigmoid outputs exceeded \(\hat{q}\), ensuring that the true label was included in the set with a specified probability. Finally, we evaluated the model via different methods (`score' and `lac'). We tested multiple confidence levels (\(\alpha = 0.2, 0.1, 0.05\)) to generate prediction intervals, ensuring the inclusion of true labels with high reliability across test datasets. Figure \ref{fig:uq2} shows that the estimated quantile values decreased as \(\alpha\) increased. A higher \(\alpha\) results in a lower quantile threshold, including more data points in the prediction set. In contrast, a lower \(\alpha\) (e.g., \(\alpha = 0.05\)) leads to a higher quantile threshold, allowing fewer but more confident predictions. Consequently, when \(\alpha\) is exceptionally high, it may lead to a low quantile value, resulting in large prediction regions. Conversely, if \(\alpha\) is too small, the quantile increases, potentially leaving no null areas of the prediction set.

\begin{figure*}[!ht]
\centering
\begin{subfigure}{0.45\linewidth} \centering
    \includegraphics[width=\linewidth]{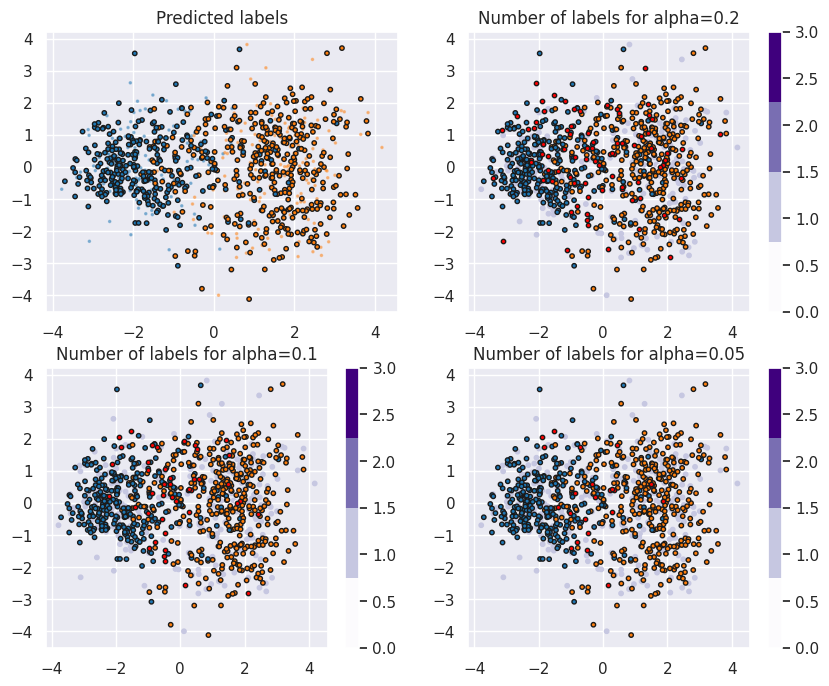}
    \caption{Visualization of the variation in prediction sets}
    \label{fig:uq4}
\end{subfigure}
\hfill
\begin{subfigure}{0.45\linewidth} \centering
    \includegraphics[width=\linewidth]{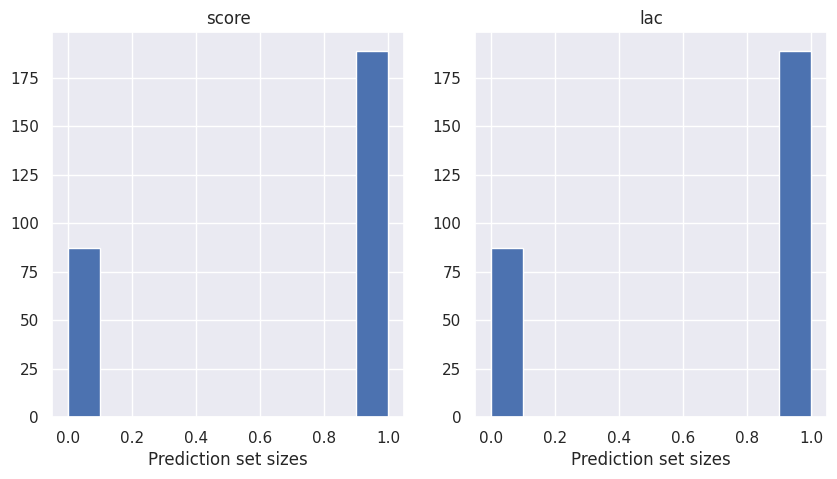}
    \caption{Prediction set sizes}
    \label{fig:uq5}
\end{subfigure}
\hfill
\begin{subfigure}{\linewidth} \centering
  \caption{Comparison of quantile, coverage score, average size of prediction sets, and empty prediction sets}
    \includegraphics[width=\linewidth]{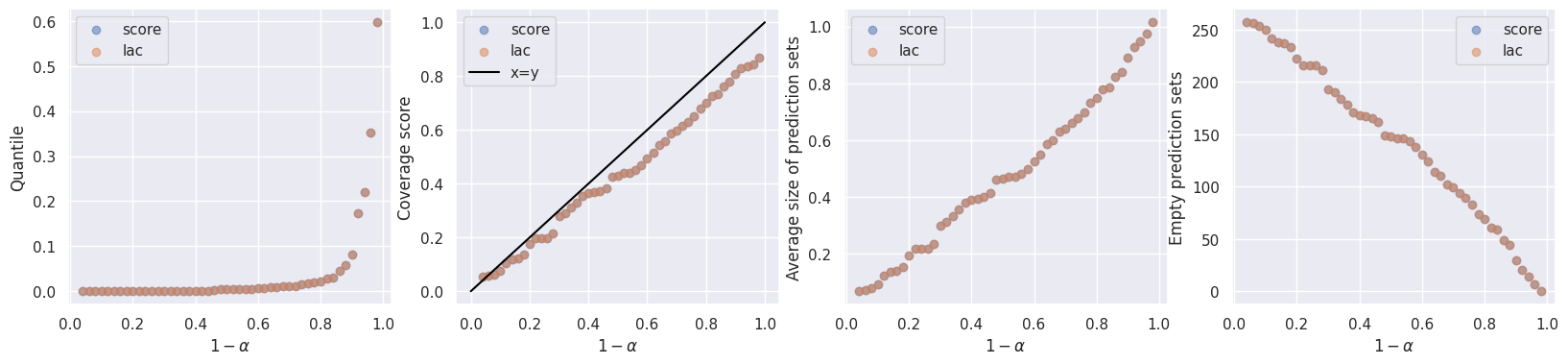}
    \label{fig:uq6}
\end{subfigure}
\caption{(a) Visualization of the differences between the prediction sets of the different values of $\alpha$. (b) Histograms of the prediction set sizes for the `score' and `lac' methods. The height of the bars at these values shows how frequently each method produces confident predictions (size = 1) versus cases of uncertainty or ambiguity (size = 0). (c) A comparison of the effective coverage, the average width of prediction sets, and the number of empty prediction sets as a function of the target coverage $1-\alpha$.}
\label{fig:uq_master2}
\end{figure*}

As shown in Figure \ref{fig:uq4}, when class coverage is low, there may be times when the model is unsure about its predictions, especially close to decision boundaries. This can cause prediction sets to be empty. As class coverage increases, these empty regions disappear, but areas of ambiguity emerge where both classes may be included in the prediction sets. This means that the selection of class coverage directly influences the prediction decisions and vice versa. A notable situation arises when predictions are based on a threshold of 0.5, as in our binary classification. In this case, each prediction set contained only a single class, eliminating ambiguous or uncertain regions. Figure \ref{fig:uq6} shows that both methods provide coverage close to the target coverage, regardless of the $\alpha$ value. The average size of prediction sets increased, and the number of empty prediction sets decreased sharply as $1-\alpha$ increased.
The histograms in Figure \ref{fig:uq5} help visualize how often each method produces single or multiple class predictions. The bars for size = 0 are insignificant, implying that the methods do not struggle with uncertainty. Similarly, a dominance of size = 1 indicates strong model confidence.

\subsection{$CO_{2}$ emissions in our research}
The development, training, and comprehensive evaluation of our proposed KACQ-DCNN model and the associated benchmark models (as detailed in Sections {\ref{sec:methods}} and {\ref{sec:results}}) were computationally intensive. These experiments were carried out on the Google Cloud Platform, specifically in the ``asia-east1'' region, with a carbon efficiency rating of 0.56 kg of $CO_{2}$ equivalent per kilowatt-hour (kWh). The research conducted $\approx400$ hours of computation using T4 hardware, which has a thermal design power (TDP) of 70W. The estimated total emissions were 15.68 kg of CO$_2$ equivalent, all fully offset by the cloud service provider. To clarify the calculations, the power consumption was determined by multiplying the power rating (70W) by the computation time (400 hours), resulting in a total energy consumption of 28 kWh. This figure was then multiplied by the region's carbon efficiency of 0.56 kg $CO_{2}$ eq/kWh to yield the total emissions of 15.68 kg $CO_{2}$ eq. If the same model was executed in the Google Cloud Platform's ``europe-west6'' region, the carbon emissions would have been lower, at 0.45 kg of CO$_2$ equivalent per kWh. 

To put the emissions of 15.68 kg of $CO_{2}$ eq into perspective, this is comparable to driving an average internal combustion engine (ICE) vehicle for $\approx63.4$ kilometers. It is equivalent to the carbon footprint of burning $\approx7.85$ kg of coal. This amount of $CO_{2}$ would be offset by the carbon sequestration capabilities of $\approx0.26$ tree seedlings over ten years.

\subsection{Interpretation of Results}
Our ablation study highlights the synergistic effects of the hybrid approach in the KACQ-DCNN model. While the ROC-AUC column shows significant variance in the ablation test, the KAN version of KACQ-DCNN 1-Layered outperforms all metrics when every component works in harmony. Model-agnostic XAI, such as LIME and SHAP, helped us precisely identify key features affecting predictions, improving the model's local and global interpretability and trustworthiness by validating feature contributions and ensuring consistent decisions across different instances. These interpretability measures provide confidence in the model's decision-making process, which is crucial for medical applications where understanding the rationale behind predictions is as important as the predictions themselves.

\subsection{Implications}
The KACQ-DCNN model demonstrates the potential of classical-quantum hybrid architectures in medical diagnostics, representing a significant advancement in computational approaches to heart disease detection. Uncertainty quantification analysis provided vital insights into the predictive performance and reliability of the model. The calibration curve indicated notable miscalibration in specific probability ranges, suggesting that recalibration methods could improve the model's probability estimates. The Mondrian approach enhanced reliability with a lower error rate and smaller prediction set sizes, leading to more accurate predictions when used for conformal prediction analysis. Additionally, the Mondrian method demonstrated higher confidence by producing single-class predictions in more cases, further validating the robustness of our approach and its potential for clinical implementation.

\subsection{Limitations}
Despite promising results, our approach faces several limitations that warrant consideration. The dependency on quantum hardware availability, which is still in the early development stages, limits the practical implementation and scalability of the KACQ-DCNN model. Quantum decoherence issues present technical challenges, as qubits can lose entanglement from heat and light, causing data loss that could affect diagnostic accuracy. The potential for incorrect qubit rotations leading to inaccurate results raises concerns about possible misdiagnoses in clinical settings. Current CQML models are limited by quantum errors, affecting reliability in sensitive applications where precision is paramount. The complexity of model training, requiring careful tuning of classical and quantum components, presents implementation challenges for widespread adoption. Furthermore, the model's sensitivity to data quality and representativeness, particularly with short and tabular datasets, could impact performance in diverse clinical environments. Finally, the dual-channel design choice may diminish the architecture's effectiveness in real-time scenarios, potentially limiting its utility in emergency medical situations that require immediate diagnostic feedback.

\subsection{Future Work}
To further improve model accuracy, scalability, and interpretability, future research directions can explore several avenues. One promising trajectory lies in leveraging quantum advantage through fault-tolerant quantum computing, which could unlock deeper optimization of dressed quantum circuits beyond current noise thresholds. Future models may integrate variational quantum kernels, allowing more expressive decision boundaries suitable for non-Euclidean data structures such as graphs or time series. Integration with symbolic AI could enable KANs to learn interpretable symbolic rules directly from data, bridging the gap between connectionist and rule-based systems. Additionally, coupling KANs with causal inference frameworks could aid in deriving explainable causal relationships from medical datasets. In terms of scalability and generalization, federated learning approaches incorporating privacy-preserving quantum federated KACQ architectures may enable decentralized yet secure training across hospital networks. Similarly, using self-supervised or contrastive learning pretraining in hybrid KAN-QML models could minimize the dependency on labeled data while preserving interpretability. Moreover, incorporating neuro-symbolic reasoning and human-in-the-loop feedback mechanisms into KACQ-DCNN could facilitate real-time clinical collaboration and decision justification, ensuring both trustworthiness and adaptability in high-stakes domains like healthcare.

\section{Conclusion}
\label{sec:conclusion}
In this study, we proposed the KACQ-DCNN model, a novel classical-quantum hybrid architecture that replaces MLP with KAN layers for heart disease diagnosis. Our comprehensive benchmarking revealed that the KAN-enhanced KACQ-DCNN 1-Layered model outperformed 34 benchmark models, including 16 CML models, 12 QNN models, and 6 hybrid models. The model achieved an accuracy of 92.03\% $\pm$ 0.19\% with strong performance across multiple metrics, including maP (92.00\% $\pm$ 0.19\%), maR (92.00\% $\pm$ 0.10\%), maF1 (92.00\% $\pm$ 0.07\%), ROC-AUC (94.77\% $\pm$ 0.43\%), MCC (83.85\% $\pm$ 0.81\%), and Cohen's kappa (83.84\% $\pm$ 0.43\%). Statistical significance was established via the Bonferroni correction ($\alpha_{\text{adjusted}} = 0.0056$), with all p-values falling well below this threshold, confirming KACQ-DCNN's superiority.

Future research should aim to incorporate more diverse and representative biomedical datasets to improve the generalizability of the KACQ-DCNN model across varying patient populations and clinical environments. Further advancements in uncertainty quantification techniques could enhance the reliability and clinical trustworthiness of predictive outcomes. Refining the model architecture with a focus on interpretability may facilitate its adoption by healthcare professionals, enabling more transparent decision support. As quantum computing technologies advance, subsequent studies may leverage next-generation quantum systems to achieve increased computational efficiency and scalability. Finally, expanding the diagnostic scope of KACQ-DCNN to include related conditions such as diabetes and sleep apnea represents a critical step toward developing a unified and comprehensive tool for multi-disease screening and diagnosis in clinical practice.

\section*{Data availability}
The datasets used in this study are publicly available on Kaggle at the following link: \\
\href{https://www.kaggle.com/datasets/fedesoriano/heart-failure-prediction}{https://www.kaggle.com/datasets/fedesoriano/heart-failure-prediction}. 


\section*{Author contributions statement}
\textbf{M.A.J.}: Conceptualization, Methodology, Data curation, Writing - Original Draft Preparation, Software, Visualization, Investigation, Validation.
\textbf{M.A.M.}: Data curation, Writing - Original Draft Preparation, Software, Visualization.
\textbf{M.F.M.}: Supervision, Reviewing, and Editing.
\textbf{Z.A.}: Advising, Reviewing, and Editing.
\textbf{N.D.}: Advising, Reviewing, and Editing.

\section*{Competing interests}
The authors have no conflict of interest, financial or otherwise, to declare relevant to this article.

\section*{Additional information}
Correspondence and requests for materials should be addressed to M.A.J.

\bibliographystyle{apalike} 
\bibliography{main}

\end{document}